\documentclass[10pt,twocolumn,letterpaper]{article}

\usepackage[pagenumbers]{iccv} 

\usepackage{times}
\usepackage{epsfig}
\usepackage{graphicx}
\usepackage{amsmath}
\usepackage{amssymb}
\usepackage{bm}
\usepackage{booktabs}
\usepackage[ruled,vlined]{algorithm2e}
\usepackage{xspace}

\newcommand{\myparagraph}[1]{\noindent\textbf{#1}}

\newcommand{\model}{PRIMAL\xspace}
\newcommand{\modelLong}{Physically Reactive and Interactive Motor model for Avatar Learning}

\definecolor{iccvblue}{rgb}{0.21,0.49,0.74}
\usepackage[pagebackref,breaklinks,colorlinks,allcolors=iccvblue]{hyperref}

\title{PRIMAL: Physically Reactive and Interactive Motor Model for Avatar Learning}

\author{
  Yan Zhang$^{1}$,\;  Yao Feng$^{1,3}$,\; Alp\'{a}r Cseke$^1$,\; Nitin Saini$^{1}$,\\ Nathan Bajandas$^{1}$,\; Nicolas Heron$^{1}$,\; Michael J. Black$^{2}$ \\
  $^1$Meshcapade,\;
  $^2$Max Planck Institute for Intelligent Systems, T\"{u}bingen\; 
  $^3$Stanford University \\ 
}

\begin{document}

\twocolumn[{%
\renewcommand\twocolumn[1][]{#1}%
\maketitle
\begin{center}
    \centering
    \captionsetup{type=figure}
    \includegraphics[width=\textwidth]{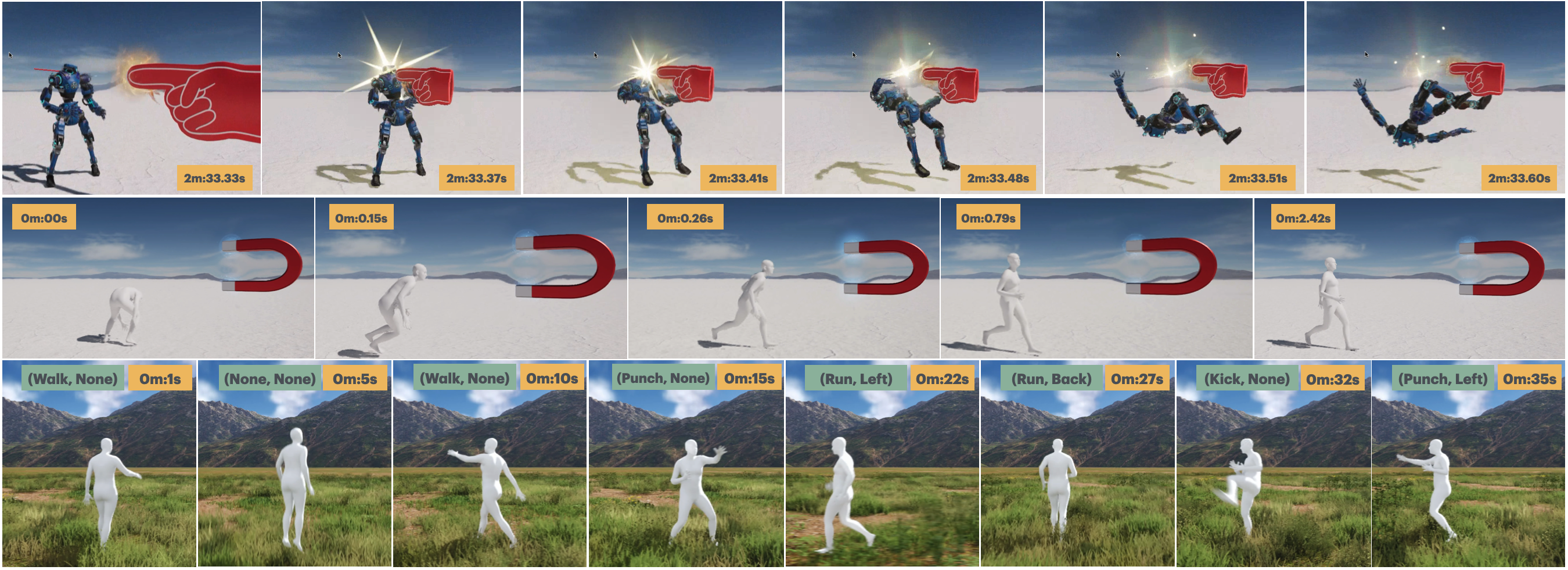}
    \captionof{figure}{\model is a novel generative real-time 3D character animation system that works in Unreal Engine.
    The avatar reacts to induced impulses promptly and naturally (top). After efficient adaptation, the avatar can be pulled to chase a ``magnet'' (middle).
    We also personalize the avatar's movements based on a tiny mocap dataset, captured by Mocapade3.0~\cite{mocapade} from cellphone videos (bottom). As a result, we can control the avatar with discrete commands and continuous signals. Without any control signal, or external perturbations, the avatar moves autonomously in the 3D space without end. \model is purely data driven; no physical simulation is used. 
    }
    \label{fig:teaser}
\end{center}%
}]

\maketitle

\begin{abstract}

We formulate the motor system of an interactive avatar as a generative motion model that can drive the body to move through 3D space in a perpetual, realistic, controllable, and responsive manner. Although human motion generation has been extensively studied, many existing methods lack the responsiveness and realism of real human movements.
Inspired by recent advances in foundation models, we propose \model, which is learned with a two-stage paradigm. 
In the pretraining stage, the model learns body movements from a large number of sub-second motion segments, providing 
a generative foundation
from which more complex motions are built. 
This training is fully unsupervised without annotations.
Given a single-frame initial state during inference, the pretrained model not only generates unbounded, realistic, and controllable motion, but also enables the avatar to be responsive to induced impulses in real time.
In the adaptation phase, we employ a novel ControlNet-like adaptor to fine-tune the base model efficiently, adapting it to new tasks such as few-shot personalized action generation and spatial target reaching. 
Evaluations show that our proposed method outperforms state-of-the-art baselines. We leverage the model to create a real-time character animation system in Unreal Engine that feels highly responsive and natural.
\footnote{Code, models, and more results are available at:\\ \url{https://yz-cnsdqz.github.io/eigenmotion/PRIMAL}}

\end{abstract}

\section{Introduction}
\label{sec:intro}

There has been rapid progress in 3D human motion generation conditioned on text or existing motion sequences \cite{tevet2023human,zhang2022motiondiffuse,jiang2024scaling,jiang2023motiongpt,zhao2023synthesizing,li2024egogen,li2024lodge,liang2024omg,li2024controllable,GRAB:2020,tripathi2024humos}.
Despite the realism of such approaches, they are not good models of how real humans behave, in that they are not interactive.
Autonomous characters need to respond in real-time to commands and environmental inputs such as external forces. 
To feel ``alive'', such an agent must be continuously on-line and reactive.
It must move on its own yet be able to respond to commands.
We introduce such an agent using a two-stage approach, in which we first train a low-level motor system from motion capture (mocap) data such that it mimics the dynamics of human motion.
This provides a platform upon which to build more complex interactive behaviors in a second stage. 
This two-stage approach provides a base ``motor system'' that implicitly models the dynamics of human motion, generalizes beyond the training data, and supports the training of more complex behaviors.

Our approach, called \model (\modelLong), models low-level motor control using an autoregressive diffusion model that is trained to capture human motion movement over short time intervals.
Without any physics simulation, we observe that this approach is able to model physically realistic movements, including ground contact, even though we never explicitly model or label contact.
This provides a data-driven approach to learning avatar motor control that appears physically realistic without physics simulation.

We make two key observations that ground our approach.
First, existing autoregressive motion models usually predict a future motion conditioned on a past motion. 
Since a motion sequence implicitly contains semantic information, this ensures the predicted motion is semantically related to the past. However, it raises the risk that the model overfits to the semantics in the training data when the training samples are limited (as they are with existing mocap datasets).
This limits generalization, resulting in a model that does not make a good foundation for general motor control.
Second, human motion is dominated by physics over short time intervals and by semantics over long ones. We use different approaches to model each of these.
We observe that physics plays a critical role in time intervals on the order of 0.5 sec, or 15 frames at 30fps. 
On longer time horizons, higher level goals dominate.

In \model, the first stage models this short-term motion: given an initial condition of the body, including its joint locations and velocities at a single time instant, we predict the motion of the next half second. 
This assumes that, over a short time horizon, the motion of the future is independent of the past.
With this assumption, we have vast amounts of motion capture data with which to train the model; every segment of 15 frames is available for training. 
We demonstrate that training our generative model in this way mimics the dynamics of human motion {\em without doing any physics simulation}.
Note that, while we do not explicitly model foot-ground contact, the model learns to produce movements with physically plausible contact.

Motivated by the representation power of diffusion models \cite{sohl2015deep,ho2020denoising,song2019generative}, the first stage is implemented as a diffusion model, which is trained on diverse half-second motion segments in a fully unsupervised manner.
The trained model not only generates realistic and autonomous motions in 3D space, but also makes it straightforward to control the avatar in real-time. 
For example, we simply reset any joint velocities to new values and the model will adapt its future motion appropriately. 
We can also drive the avatar to reach desired speeds and directions, using a classifier-based guidance (CBG)~\cite{dhariwal2021diffusion} approach, for which the CBG gradients are analytically derived.

High-level and long-term behaviors can be composed as a time sequence of consecutive atomic actions.
This decouples the learning of movements and semantics, enabling us to learn them individually in a two-stage paradigm with a pretraining stage (motor control) and the adaptation stage (behavior generation).
Specifically, we propose a generic adaptation approach based on ControlNet~\cite{zhang2023adding} in the second stage.
Here, we consider two types of adaptation. One focuses on spatial target reaching and the other on few-shot semantic action generation.
To improve the effectiveness of the control signal, we follow~\cite{liang2024omg} to add the control embedding to individual blocks of the transformer.

We demonstrate the benefits of our interactive avatar with several applications.
First, we create an interactive 3D environment in Unreal Engine, in which the agent exhibits perpetual motion (see Fig.~\ref{fig:teaser}).
The agent's movement is controlled with standard game controllers to move in any direction at varied speeds. 
Unlike traditional game technology, however, all the motions are uniquely generated in real time based on the commands.
We introduce external forces (``pulling'' with a ``magnet'' or an impulsive force that ``zaps'' the character) that dynamically change the agent's behavior (see {\bf Supplemental Video}). 
The agent reacts promptly to interventions in a natural way, without any explicit physics-based training, showing that the first stage learns a generic motor model. 
Second, we demonstrate how our approach enables training a person/style-specific model using a tiny amount of motion capture data captured with a cell phone.
The results illustrate how pre-training a foundational movement stage provides a basic ``motor system'' upon which to build complex behaviors.

In summary, \model produces naturalistic human motion common to methods trained on mocap data together with the physical plausibility and generalization of physics-based methods.
To build a motor system for interactive avatars, we use an autoregressive diffusion model with a two-stage pretrain-and-adaptation learning paradigm.
This first stage trains the model to capture human motion over short time scales, providing a reactive model that runs in real time. 
The second stage uses an efficient and generic approach to adapt the base model to different avatar-related tasks.
The interactive nature of \model effectively ``breathes life'' into an avatar, making it appropriate for applications in gaming, AR/VR, controllable video generation, and interactive agents.

\section{Related Work}
\label{sec:related}

\paragraph{Offline motion generation.}
Enabled by large mocap datasets \cite{AMASS:ICCV:2019,guo2022generating,BABEL:CVPR:2021}, generative motion models have advanced rapidly, including methods focused on
text-to-motion \cite{tevet2023human,zhang2022motiondiffuse,jiang2023motiongpt,liang2024omg,huang2024como,dai2024motionlcm},
human-scene/object interactions \cite{jiang2024scaling,zhao2023synthesizing,li2024egogen,li2023object,GRAB:2020,wu2024human,hassan2023synthesizing,hassan_samp_2021,araujo2023circle,taheri2021goal,wu2022saga}, 
spatial control \cite{ling2020character,zhang2022wanderings,karunratanakul2023gmd,xie2024omnicontrol,diomataris2024wandr,rempe2023trace,wang2024pacer+}, dance generation \cite{li2024lodge,tseng2023edge,li2024interdance}, and more.
Specifically, diffusion models \cite{sohl2015deep,song2019generative,ho2020denoising} are a promising approach for motion generation, due to their representation power and controllability. 
For example, MDM~\cite{tevet2023human} proposes a diffusion model with the transformer~\cite{vaswani2017attention} network, which is trained with text-motion paired data \cite{guo2022generating} and is able to generate high-fidelity motions given text descriptions.
OmniControl~\cite{xie2024omnicontrol} leverages ControlNet~\cite{zhang2023adding} to create a new adaptor, enabling spatial control of text-to-motion.
MotionLCM~\cite{dai2024motionlcm} proposes a consistency model for real-time text-to-motion, and implements autoregressive generation as a spatial control task.
Most of these approaches are offline methods and have motion length limitations, making them inappropriate for interactive avatar control.

More similar to our approach is OMG~\cite{liang2024omg}, which proposes a pretrain-then-finetune strategy for text-to-motion. Its pretraining phase trains foundation models with up to 1 billion parameters only with unlabeled mocap data, and the fine-tune phase leverages a ControlNet-like adaptor with a mixture-of-experts. 
In contrast to our approach, OMG is offline and their pretraining process focuses on generating motions up to 300 frames long, much longer than our short time horizon. 
Unlike OMG, our model formulation is autoregressive, which not only generates unbounded and controllable motion in real time, but also makes the avatar reactive to external perturbations.

\myparagraph{Physics-based motion generation.}
A key problem of the data-driven methods above is that they often lack physical plausibility; e.g.~feet slide, float, or penetrate the floor, and impacts are not realistic.
To address this, many methods have been developed that exploit physical simulation
\cite{liu2024physreaction,peng2021amp,rempe2023trace,hassan2023synthesizing,wang2024pacer+,ren2023insactor}.
Hassan et al.~\cite{hassan2023synthesizing} employ reinforcement learning (RL) and adversarial imitation learning to produce life-like and physically plausible human-scene interactions.
PACER++~\cite{wang2024pacer+} generates physically plausible pedestrian locomotion in driving scenarios with fine-grained body part control.
Simulation can also be efficiently combined with generative models.
For example, PhysDiff~\cite{yuan2023physdiff} projects the generated motion into a physically plausible motion space via physical simulation, which is applied within the reverse diffusion process.
CLoSD~\cite{tevet2024closd} uses an efficient autoregressive diffusion model, DiP, for planning and applies simulation for tracking, generating unbounded physically plausible motions in 3D.
InsActor~\cite{ren2023insactor} proposes an hierarchical approach to generate motions given instructions, which contains diffusion-based planning and simulation-based motion tracking.
An advantage of simulation approaches is that they can generalize to adapt to external forces.
However, the resulting motions are usually not yet as natural as those from diffusion models. 
Additionally, simulations tend to learn policies of a narrow set of motor skills, rather than a generic human motion control system.

\myparagraph{Autoregressive unbounded motion generation.}
Autogregressive models are trained to generate future frames conditioned on past frames. 
GAMMA~\cite{zhang2022wanderings} is a VAE~\cite{kingma2013auto} model that generates motion primitives recursively and leverages RL to train a policy for spatial target reaching.
WANDR~\cite{diomataris2024wandr} is a conditional VAE that generates  motions to reach a 3D target with the wrist joint.
DIMOS~\cite{zhao2023synthesizing} uses scene-aware policies for motion-primitive generation, enabling the synthesis of continuous human-scene interactions.
Recently, diffusion models have been used for autoregressive motion modeling.
For example, CAMDM~\cite{camdm} exploits a diffusion model that treats motion style, spatial control, and motion history as separate tokens, producing unbounded motions for real-time character animation.
Jiang et al.~\cite{jiang2024autonomous} propose a diffusion model that is conditioned on the past 2 frames, as well as text, scene, etc., to generate human-scene interactions.
AMDM~\cite{shi2024amdm} uses an autoregressive diffusion model with an MLP denoiser. 
In addition to generating random motions, policy networks are trained with RL for motion control in various tasks.
The DiP model proposed in CLoSD~\cite{tevet2024closd} is conditioned on the motion history of 20 frames and text tokens, and generates a 40-frame motion in the future. 
DART~\cite{Zhao_DART_2024} uses an online generation method that has an autoregressive VAE tokenizer and a latent autoregressive diffusion model, modeling motion primitives and performing high-fidelity text-to-motion in real time. Latent space optimization and RL policies are used for motion control on the pre-trained model. 

Compared to existing autoregressive diffusion models that generate $N$ future frames based on $M$ past frames, our model formulation 
generates motions based on an initial state of \textit{a single frame} that includes joint positions and velocities.
With this form, \model learns human movement from unstructured and unlabeled mocap data directly, without the support of physics or annotations. 
Despite the lack of explicit physics training, we find that visually plausible physical effects are present in the generated motions.
In addition, neither motion tokenizers nor scheduled sampling~\cite{bengio2015scheduled} schemes are necessary in our method, leading to a simple and straightforward training workflow.

\section{Autoregressive Diffusion Motion Models}
\label{sec:method}

\subsection{Preliminaries}

\paragraph{SMPL-X~\cite{SMPL-X:2019}.}
\label{sec:smplx}
We employ the SMPL-X gender-neutral body model and denote it as $\mathcal{M}(\beta, r, \varphi, \theta)$, in which $\beta\in\mathbb{R}^{16}$ represents the body shape parameters, $r\in\mathbb{R}^3$ is the root translation, $\varphi \in \mathbb{R}^6$ and $\theta\in\mathbb{R}^{126}$ are the root and joint rotations w.r.t.~the continuous 6D representation~\cite{zhou2019continuity}, respectively.
Since we focus on the body motion without facial expressions or hand articulations, we extract the 22 joints of the body skeleton and denote their locations and velocities as $J\in\mathbb{R}^{66}$ and $\dot{J}\in\mathbb{R}^{66}$, respectively.

\paragraph{Denoising diffusion probabilistic models (DDPM)~\cite{ho2020denoising}.}
Diffusion models are widely used for tasks like motion generation.
In this work, we follow the DDPM formulation~\cite{ho2020denoising}.
Provided a clean sample $\mathbf{X}^0 \sim p(\mathbf{X}^0)$, the forward diffusion process gradually adds Gaussian noise to the sample in $T$ steps, and produces a sequence of samples at different noise levels, i.e. $\{\mathbf{X}^1, ...,\mathbf{X}^t,...,\mathbf{X}^T\}$.
Given $q(\mathbf{X}^t | \mathbf{X}^{t-1})=\mathcal{N}\left(\sqrt{1-b_t}X^{t-1}, b_t\mathbf{I}\right)$ with $\{b_t \in (0,1)\}_{t=1}^T$ being a set of noise variance scheduler variables, it has
\begin{equation}
    \mathbf{X}^t = \sqrt{\bar{\alpha}_t}  \mathbf{X}^0+ \sqrt{1-\bar{\alpha}_t} \mathbf{\epsilon},
\end{equation}
in which $\alpha_t = 1-b_t$, $\bar{\alpha}_t = \prod_{t'=1}^t \alpha_{t'}$, and $\mathbf{\epsilon}\sim\mathcal{N}(0,\mathbf{I})$.

The reverse diffusion process gradually removes noise from $\mathbf{X}_t$ to recover the original clean sample $\mathbf{X}_0$. 
Following~\cite{ramesh2022hierarchical,tevet2023human}, we use a neural network to predict the clean sample in this work, i.e.~$\hat{\mathbf{X}}^0=G(t, \mathbf{X}^t, \mathbf{c})$, which takes the diffusion time step, the noisy sample, and a condition as input. 
This network is trained by minimizing a {\em simple} loss~\cite{ho2020denoising}, given by $\mathcal{L}_{\text{simple}} = $
\begin{equation}
\label{eq:simpleloss}
    \mathbb{E}_{\mathbf{X}^0 \sim p(\mathbf{X}^0), t\sim U[1,T]} \left[ \|\mathbf{X}^0 - G(t,\mathbf{X}^t, \mathbf{c}) \|^2 \right].
\end{equation}

During test-time, the reverse diffusion process can be controlled via classifier-based guidance (CBG)~\cite{dhariwal2021diffusion} and classifier-free guidance (CFG)~\cite{ho2022classifier}.
With a pre-trained diffusion model and the guidance objective functions, CBG computes the gradients of the objectives at each denoising step, and hence does not support real-time applications in general.
On the other hand, CFG trains the conditional and unconditional diffusion models jointly, and guides the reverse diffusion process via weighted averaging, which is usually much more efficient than CBG.
With $G(t,\mathbf{X}^t, \emptyset)$ being the unconditional model, CFG is performed at each denoising step by
\begin{equation}
    \label{eq:cfg}
    \hat{X}^0 = G(t,\mathbf{X}^t, \emptyset) + \gamma (G(t,\mathbf{X}^t, \mathbf{c})-G(t,\mathbf{X}^t, \emptyset)),
\end{equation}
in which $\gamma \in \mathbb{R}^+$ is a hyper-parameter. 
A larger value of $\gamma$ will increase the influence of the condition. However, invalid results can be produced if it is too large.

\subsection{Motion Representation}
\label{sec:method:motion_repr}

We represent an $N$-frame motion segment as a time sequence of states, 
${\bm X} = \{{\bm x}_0, {\bm x}_i,...,{\bm x}_{N-1} \}_{i=0}^{N-1}$.
Each state has ${\bm x}_i = (r, \varphi, \theta, J, \dot{J}) \in \mathbb{R}^{267}$ based on SMPL-X. 
During training, we use the ground truth body shape ($\beta$).
During testing, we can either leverage a ground truth $\beta$ from the test set, or draw it from $N(0,\mathbf{I})$, in order to compute the initial state ${\bm x}_0$. During motion generation, $\beta$ is fixed.

Each motion segment spans 0.5 seconds in this work.
This time duration corresponds to the average hang time of human jumps\footnote{Hang time \cite{PMID:522639} describes how long someone can stay in the air under their own power and is a function of gravity and how high a person can jump.} or nearly half of a walking cycle. This is a time frame during which human motion is dominated more by physical forces than by intentional control.

Following the canonical coordinate system proposed by~\cite{zhang2021we,zhang2022wanderings}, the body-centric canonical coordinate is located at the pelvis of the first frame, with X,Y, and Z pointing left, up, and forward, respectively.
The neural network is trained with canonicalized motion segments.

\subsection{Neural Networks}
\label{sec:method_net}

\begin{figure}
    \centering
    \includegraphics[width=\linewidth]{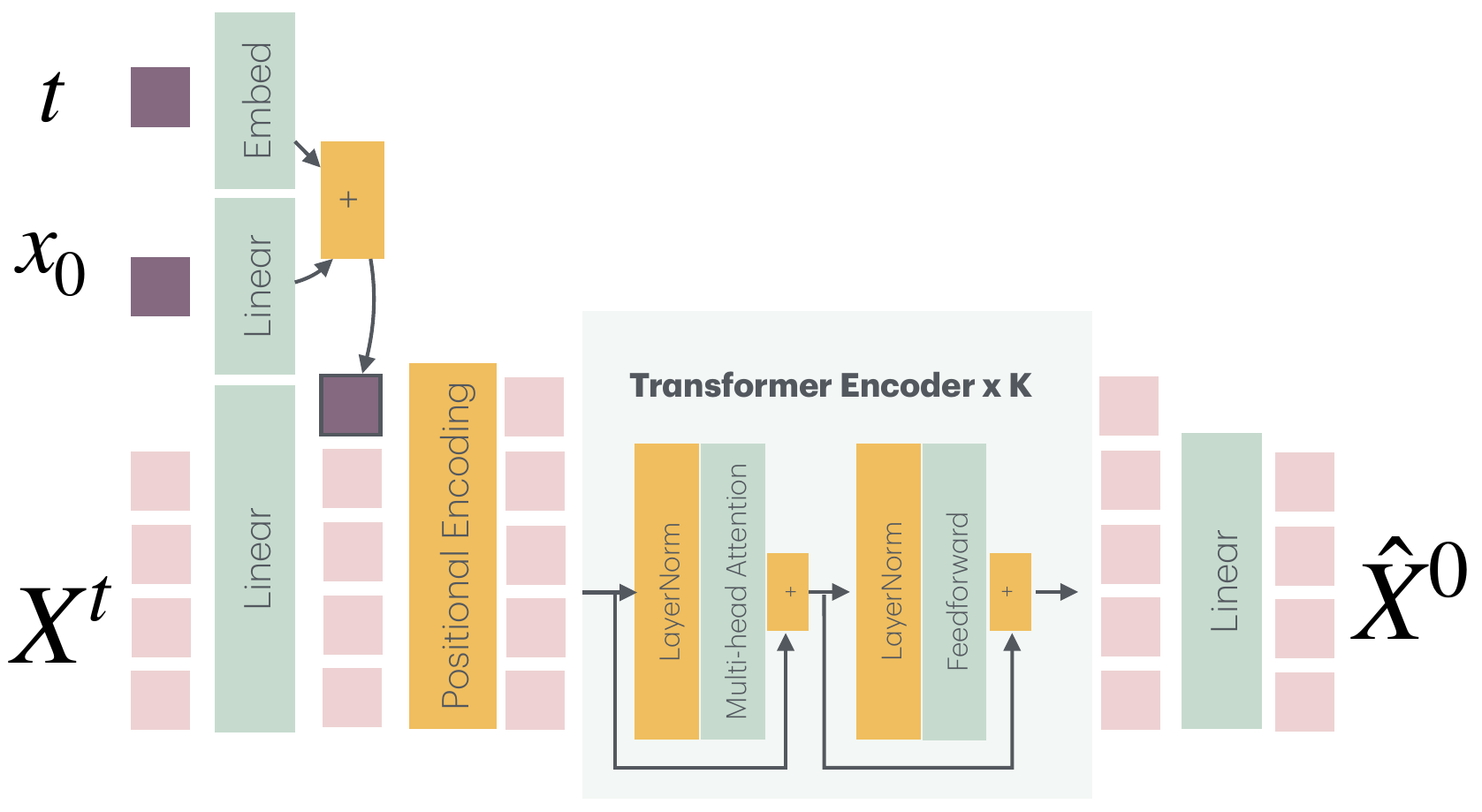}
    \caption{Illustration of our network $\hat{\bm{X}}^0 = G(t, {\bm{X}}^t, \bm{x}_0)$. 
    The tea-green layers contain trainable parameters and the orange blocks are non-learnable operations. 
    The pink squares denote the tokens at individual frames.}
    \label{fig:base_models}
\end{figure}

\paragraph{Architecture.}
Following e.g.~\cite{camdm,shi2024amdm,tevet2024closd,Zhao_DART_2024}, in this paper the autoregressive diffusion model is defined as a diffusion model that generates $N$ future frames based on $M$ past frames. 
We formulate our denoising network as $G(t, \mathbf{X}^t, \bm{x}_0)$ with transformers~\cite{vaswani2017attention,peebles2023scalable} and illustrate the architecture in Fig.~\ref{fig:base_models}.
It leverages in-context conditioning, in which the diffusion time embedding and the initial state embedding are first added and then concatenated with the embedding of the noisy motion segment.
SiLU~\cite{hendrycks2016gaussian,elfwing2018sigmoid,swish} is used in the feed-forward layers.

\paragraph{Training.}
Unlike existing autoregressive diffusion models, e.g.~\cite{Zhao_DART_2024,tevet2024closd,camdm}, that leverage annotated training data, our model focuses on modeling the body movements and is trained only with unlabeled short mocap segments. Therefore, the data limitation issue is effectively overcome, since we can collect a huge amount of segments from AMASS \cite{AMASS:ICCV:2019}.
In addition, our motion model is directly trained in the motion space, instead of the latent space as in~\cite{jiang2023motiongpt,Zhao_DART_2024}. No separate tokenizers are trained in our method.

Our training loss is given by
\begin{equation}
    \label{eq:our_loss}
    \mathcal{L} = \mathcal{L}_{\text{simple}} + \gamma_1 \mathcal{L}_{\mathit{FK}} + \gamma_2 \mathcal{L}_{\mathit{FKV}},
\end{equation}
which contains a DDPM \textit{simple} term (see Eq.~\eqref{eq:simpleloss}), two forward kinematics terms on the joint locations and velocities, respectively, and two weights $\gamma_1$ and $\gamma_2$. 
The two forward kinematics losses are given by
\begin{equation}
    \mathcal{L}_{\mathit{FK}} = \frac{1}{N}\sum_{i=0}^{N-1} \|\mathcal{M}(\beta, \hat{r}_i, \hat{\varphi}_i, \hat{\theta}_i) - J_i\|^2_2 ,
\end{equation}
and 
\begin{equation}
    \mathcal{L}_{\mathit{FKV}} = \frac{1}{N}\sum_{i=0}^{N-1} \|\dot{\mathcal{M}}(\beta, \hat{r}_i, \hat{\varphi}_i, \hat{\theta}_i) - \dot{J}_i\|^2_2 ,
\end{equation}
in which the predicted motion state at the $i$-th frame is denoted by $\hat{\bm{x}}_i=(\hat{r}_i, \hat{\varphi}_i, \hat{\theta}_i, \hat{J}_i, \hat{\dot{J}}_i)$. The ground truth body shape $\beta$ is used.

Our model is trained in a fully unsupervised manner, and does not use foot contact labels as in~\cite{tevet2023human,zhang2024rohm}.
We think the contact information, which is usually defined by the horizontal velocities of the foot joints, can be automatically learned by minimizing the $\mathcal{L}_{\mathit{FKV}}$ term.
Incorrect annotations can also degrade model performance.
Moreover, we find that our model is self-stabilizing after being trained.
Therefore, scheduled sampling~\cite{bengio2015scheduled}, which can substantially reduce the training speed, is not employed in our training phase.
Compared to existing methods such as~\cite{ling2020character,zhang2022wanderings,zhao2023synthesizing,Zhao_DART_2024,shi2024amdm,tevet2023human,tevet2024closd,camdm}, our training pipeline is simple and straightforward.

\subsection{Inference}
\label{sec:method_inference}
Our motion model learns the distribution of $p({\bm X} | {\bm x}_0)$. 
Given an initial state, the model generates a motion segment starting with this initial state, and  works repeatedly to produce an arbitrarily long motion sequence.
The overall inference approach is summarized in Alg.~\ref{alg:inference} in Sup.~Mat.

In addition, we propose three efficient and effective test-time processing approaches to refine the motion on the fly, i.e. \textit{joint re-projection}, \textit{snapping to the ground}, and \textit{inertialization blending}~\cite{inertialization,inertialization_d}; see Sup.~Mat.~for details.

\subsubsection{Motion Control with Induced Impulses}
The model generalizes outside the training data, exhibiting behavior typically seen only with physics simulation.
Specifically, we perturb the avatar with novel external impulses by changing the velocity at one or more joints.
Using this, we can control the avatar with impulses to generate certain actions in a \textit{principled manner}. 
For example, we can generate kicks by giving an upward impulse to the joints on the leg, or generate runs by pushing the avatars from the back.

Impulse induction is implemented by perturbing the initial velocities before generating a new segment. 
We find that such impulses can be defined to reliably generate specific actions; see Sup.~Mat.~Tab.~\ref{tab:five_actions}.
Future work should explore the automated discovery of impulses for controlling behavior.

\subsubsection{Motion Control with Classifier-based Guidance}
\label{sec:method_cbg}
Another basic level of controllability needed for an interactive avatar is the ability to move in a given direction at a given speed.
To enable that, we propose two intuitive classifier-based guidance (\textbf{CBG}) loss terms to control the movement and the facing direction of the avatar.

The movement guidance loss is based on the average velocity of all 22 joints in the half-second time horizon, which is given by
\begin{equation}
    \mathcal{L}_{\mathit{move}} = \left\| \left(\frac{1}{22N}\sum_{i=0}^{N-1}\sum_{j=1}^{22}\hat{\dot{J}} \right) - \bm{v}_{\mathit{goal}} \right\|^2,
\end{equation}
in which $\bm{v}_{goal} \in \mathbb{R}^3$ denotes the target mean speed.

The facing guidance loss is based on the last frame of the motion segment.
Given the avatar's normalized facing directional vector ${\bm z}$ and a target facing unit vector $\bm{r}_{goal}$, the guidance loss is given by
\begin{equation}
    \mathcal{L}_{\mathit{facing}} = \left\| {\bm z} - \bm{r}_{\mathit{goal}} \right\|^2,
\end{equation}
in which ${\bm z}$ is computed from the predicted joint locations.

In order to obtain real-time performance, we derive the analytical gradients of these terms instead of using auto-differentiation.
Note that we control motions via manipulating $J$ or $\dot{J}$ in the motion state $\bm{x}_i$. The SMPL-X parameters, i.e. $(r, \varphi, \theta)$, are not directly controlled.

\section{Model Adaptation}
\label{sec:method:adaptation}

\begin{figure}
    \centering
    \includegraphics[width=\linewidth]{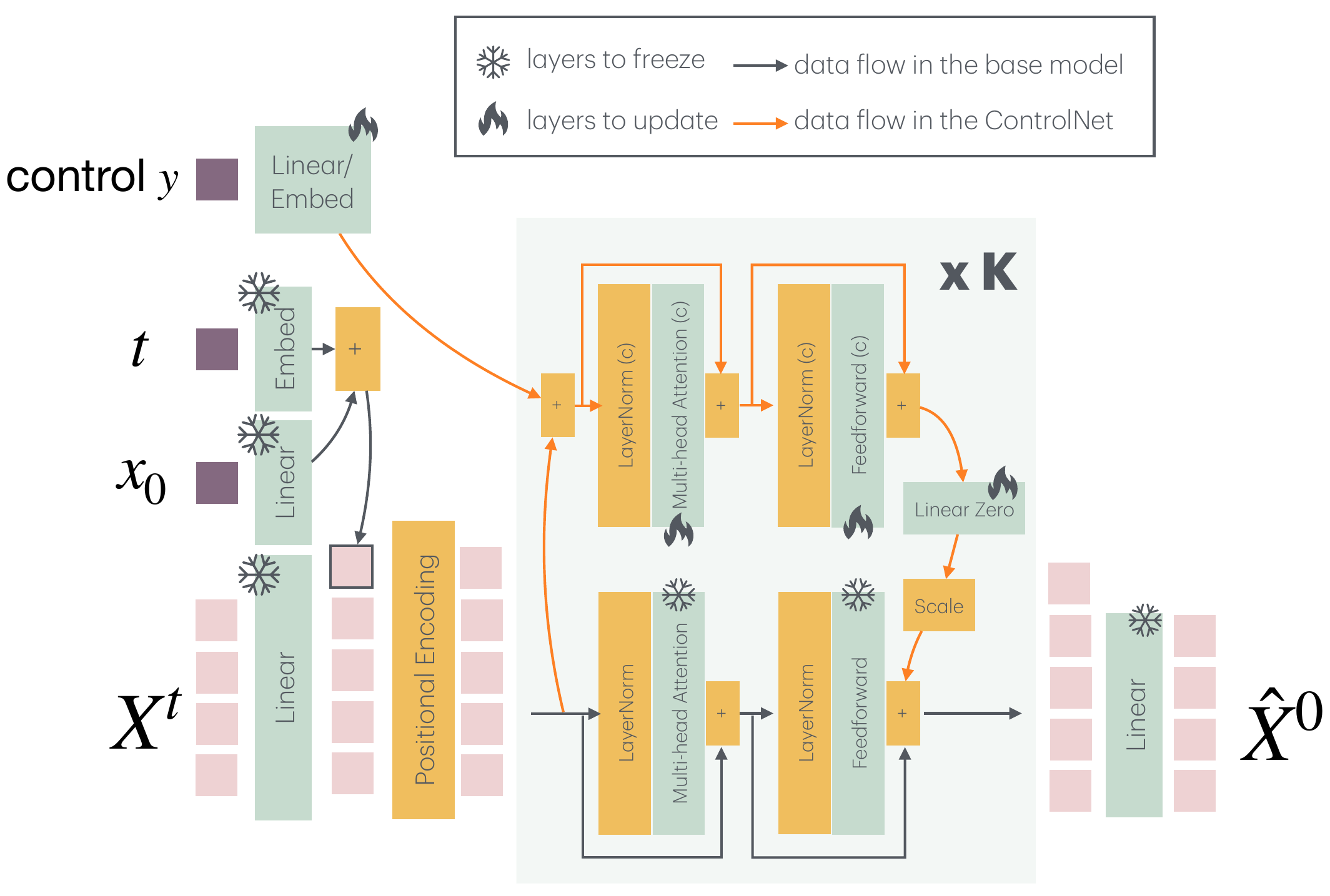}
    \caption{Illustration of our adaptation method. Given a base model, we introduce a ControlNet-like adaptor at each individual transformer block. The control signal $y$ is a generic notation. We adapt the base model for semantic action generation and spatial target reaching. The {scale} operation manipulates the impact of the control signal, corresponding to $\gamma$ in Eq.~\eqref{eq:cfg} of CFG.
    }
    \label{fig:net_adaptation}
\end{figure}

Since it takes 3-4 days to train our autoregressive diffusion model to obtain high-quality results, we regard it as a base model for task-specific adaptation; this reduces the costs of data and computation.
Therefore, we do not need to train from scratch to solve new tasks.
The effectiveness of this approach has been widely reported in recent advances of foundation models~\cite{OpenAI_ChatGPT,touvron2023llama,hu2022lora,rombach2022high,zhang2023adding,nvidia2025cosmosworldfoundationmodel}.

In this work, we take advantage of ControlNet~\cite{zhang2023adding} and propose a generic approach for task adaptation, which is illustrated by Fig.~\ref{fig:net_adaptation}. For each adaptor, the control embedding and the output from the previous transformer block are added first and then fed into a trainable copy of the current transformer block.
The adaptation process only updates the parameters in the adaptors, while keeping the base model parameters frozen. The training loss is identical to Sec.~\ref{sec:method_net}.

This architecture is different from the ControlNets applied in~\cite{xie2024omnicontrol,zhong2024smoodi}, in which the control embedding is just input once to the adaptor.
Our transformer block-wise adaptation is similar to OMG~\cite{liang2024omg}.
However, their mixture-of-experts module is specifically designed to fuse the text tokens via cross attention, whereas ours is a generic framework that fuses a generic control embedding via addition.

\paragraph{Few-shot semantic action generation.}
Given a small set of personalized mocap sequences with action labels, we adapt our base model to generate actions given the labels.
In this case, the control signal $y$ shown in Fig.~\ref{fig:net_adaptation} is an integer action label, and an embedding layer maps this label to a latent vector.
After adaptation, our model is able to generate infinite personalized movements based on the action labels, and natural transitions between actions, beyond what this small dataset provides.
We can control the impact of the action embedding via classifier-free guidance (CFG).
Simultaneously, we can also perturb the joint velocities to generate natural reactions, and control the movement continuously via the CBG method proposed in Sec.~\ref{sec:method_cbg}.

\paragraph{Spatial target reaching.}
In this scenario, the control signal $y$ becomes a 2D target location for the avatar to reach. 
During finetuning, the pelvis XZ coordinate of the last frame of the canonicalized motion segment is treated as the target. 
Following~\cite{diomataris2024wandr,taheri2021goal}, we re-scale the target location via $2 \cdot (1- e^{-\|\bm{x}\|_2}) \cdot \frac{\bm{x}}{\|\bm{x}\|_2}$, in order to make the model generalize to targets in extreme scenarios.
In addition, we use a linear layer to map the scaled target location to a latent variable.
After training, target-reaching behavior can be synthesized by specifying the target locations at test time. 
Since the motion is generated autoregressively, one can control the avatar's motion direction by controlling the target.

\section{Experiments}
\label{sec:experiment}

\subsection{Evaluations on Body Motion Learning}
\label{sec:exp:learning_dynamics}
A model that learns human body motion well should have two capabilities:
1) It can generate high-fidelity, realistic motions, given an arbitrary initial state;
2) The model's responses to inputs should be natural and human-like.
Thus, we have two sets of experiments to measure motion realism and  responsiveness, respectively.
Below, we denote our \model architecture in Fig.~\ref{fig:base_models} as \textit{InContext}.
In addition, we compare it with another architecture that exploits adaptive layer normalization~\cite{perez2018film,peebles2023scalable}, denoted as \textit{AdaLN}, which is illustrated in Sup.~Mat.

\subsubsection{Motion Realism}
\label{sec:exp:motion_realism}

\paragraph{Datasets and metrics.}
We draw 80 and 160 initial states from the HumanEva~\cite{sigal2010humaneva} and SFU~\cite{mocap_sfu} subsets of AMASS, respectively.
We use the gender-neutral SMPL-X body and replace the ground truth shapes with random samples from $\mathcal{N}(0,I)$, testing generalization to unseen identities. From an initial state, we generate a 240-frame motion, ie.~8s.

We evaluate the motion realism w.r.t.~the following terms.
1) \textit{Averaged skating ratio (ASR)}:
Following~\cite{zhang2021we,zhang2022wanderings,zhao2023synthesizing}, we define foot skating as occurring when both foot joints are in contact with the ground (within a tolerance of 5cm) and their horizontal velocities are both above 15cm/s. We report the averaged ratio of skating frames.
2) \textit{Averaged non-collision ratio (ANCR)}:
Similar to~\citep{yuan2023physdiff}, we compute the lowest body joint at each frame, and define a collision when the lowest body joint is deeper than 5cm below the ground plane.
3) \textit{Averaged neural Riemannian distance (ARD)}:
We employ NRDF~\cite{he24nrdf} to measure the distance of the generated body pose to the pretrained natural pose manifold of AMASS.
A smaller value indicates that the generated pose is more natural.
4) \textit{Baseline preference ratio (BPR)}:
For each method we render 50 videos and perform perceptual studies to visually compare \textit{InContext} with individual baselines; we report the percentage of participants who prefer the baseline results. 
See Sup.~Mat. for more details.

\paragraph{Comparisons to state-of-the-art.}
We compare our methods with two groups of baselines.
The \textit{off-the-shelf} group includes DiP~\cite{tevet2024closd}, CLoSD~\cite{tevet2024closd} and MotionLCM~\cite{dai2024motionlcm}.
The \textit{``our implemented''} group in Tab.~\ref{tab:sota} includes versions of DiP and DART~\cite{Zhao_DART_2024} where the motion representations and pretraining strategies are identical to ours.
For details of how we train and fairly compare the methods, see Sup.~Mat.

\begin{figure}
    \centering
    \includegraphics[width=\linewidth]{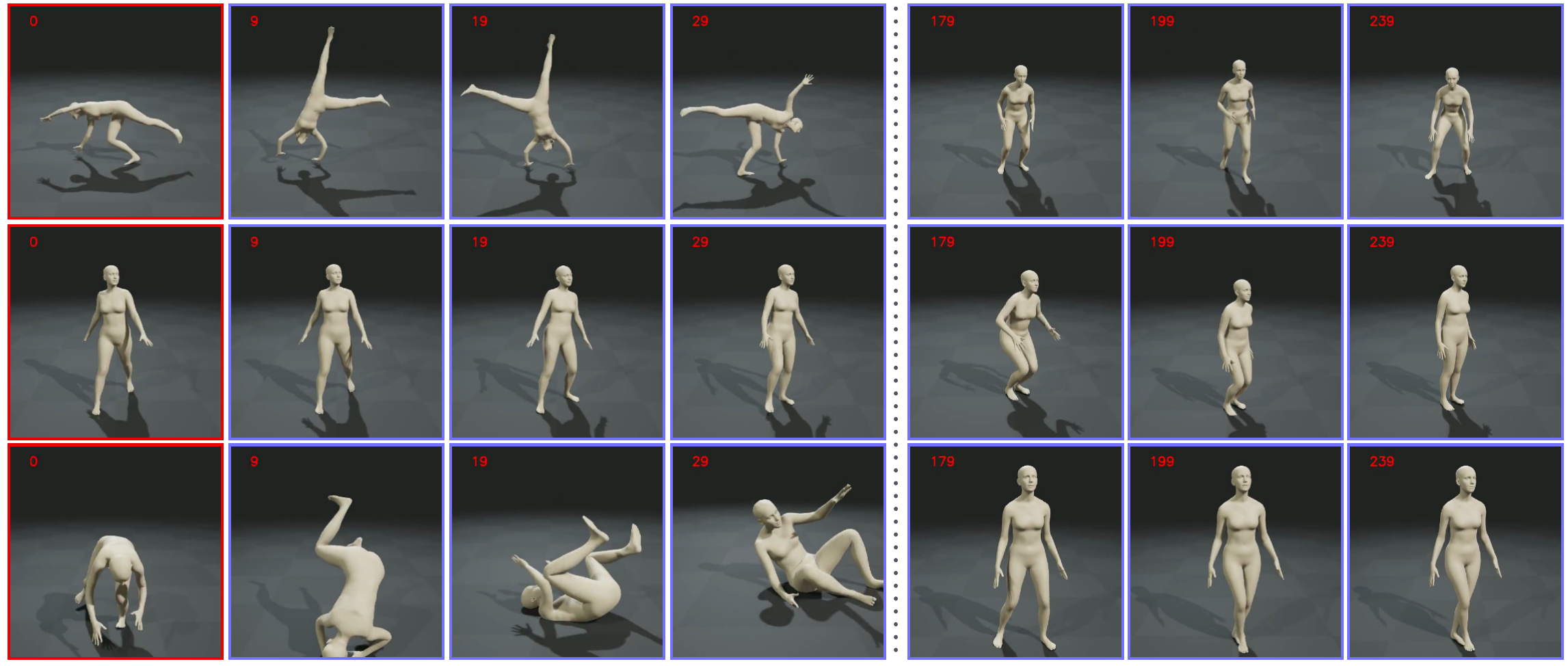}
    \caption{Illustration of three random motion generations (one per row). Given different \textcolor{red}{initial states}, the model is able to generate high-fidelity \textcolor{blue}{future motions} of arbitrary lengths in real time. The numbers denote the frame indices. See \textbf{Supplemental Video}.}
    \label{fig:vis_randommogen}
\end{figure}

Table \ref{tab:sota} shows that all methods avoid ground collision. 
Our \model-\textit{InContext} model outperforms data-driven baselines w.r.t.~foot skating, despite not explicitly training for this.
Note that \model directly outputs SMPL-X parameters, whereas DiP and CloSD need to fit SMPL-X to joint locations. SMPL-X is needed for the ARD and BPR scores and the fit step could negatively impact their scores.
Note that \textit{InContext} is visually preferred to all baselines (ie.~low BPR) suggesting the realism of the generated motions.
Figure~\ref{fig:vis_randommogen} shows generated motions given different initial states;
see the {\bf Supplemental Video}.

Furthermore, we can see that \textit{InContext} outperforms \textit{AdaLN} w.r.t.~motion realism. We also find \textit{InContext} runs slightly faster in our trials.
Therefore, we will only use \textit{InContext} in the following experiments.
More details and ablation studies are in Sup.~Mat.

\begin{table}
    \centering
    \footnotesize
    \begin{tabular}{lcccc}
    \toprule
         \textit{Methods} & \textit{ASR} $\downarrow$ & \textit{ANCR}  $\uparrow$ & \textit{ARD} $\times 10^{-3} \downarrow$ & BPR $\downarrow$\\
         \midrule
         \multicolumn{5}{c}{\textit{off-the-shelf}}\\
         DiP~\cite{tevet2024closd} & 0.067 & 1.0 & 7.373 & 20.81\% \\
         CLoSD~\cite{tevet2024closd} & \textbf{0.019} & 1.0 & 1.994 & 9.17\% \\
         MotionLCM~\cite{dai2024motionlcm} & 0.149 & 1.0 & - & - \\
         \midrule
         \multicolumn{5}{c}{\textit{our implemented}}\\
         DART~\cite{Zhao_DART_2024}  & 0.072 & 1.0 & 0.077 & 34.88\% \\
         DiP~\cite{tevet2024closd} & 0.065 & 1.0 & 0.174 & 41.15\% \\
         \midrule
         \multicolumn{5}{c}{\textit{our methods}}\\
         InContext (ours) & {0.024} & 1.0 & \textbf{0.074} & - \\
         AdaLN (ours) & 0.036 & 1.0 & 0.147 & - \\
         \bottomrule
    \end{tabular}
    \caption{Motion Realism: Comparison with SOTA. Best results are highlighted in boldface. We only compare ``InContext'' with baselines for the perceptual study (BPR). See text.
    }
    \label{tab:sota}
\end{table}

\subsubsection{Motion Responsiveness}
\label{sec:exp:principled}

The ability to respond to external input is necessary for an avatar to feel ``alive'' (see Fig.~\ref{fig:teaser}).
We draw 45 random initial states from SFU~\cite{mocap_sfu}, and replace the ground truth body shape with a random value as before.
We compare our \textit{InContext} method with our implemented DiP and DART diffusion models, and \model-\textit{InContext-8f}, which has an 8-frame time horizon as in DART.

\begin{table}
    \centering
 \resizebox{\columnwidth}{!}{
   \begin{tabular}{lcccc}
    \toprule
         \textit{Methods} & \textit{ASR} $\downarrow$ & \textit{ANCR}  $\uparrow$ & \textit{ARD} $\times 10^{-3} \downarrow$ & BPR $\downarrow$\\
         \midrule
         DART diffusion~\cite{Zhao_DART_2024} & 0.159 & 0.961 & 2.259 & 21.29\%  \\
         DiP~\cite{tevet2024closd} & 0.126 & 0.995 & 0.645 & 10.50\% \\
         \midrule
         InContext-8f & 0.172 & 0.995 & 7.903 & 30.66\% \\
         InContext & \textbf{0.088} & \textbf{1.0} & \textbf{0.511} & -\\
         \bottomrule
    \end{tabular}}
    \caption{Results of reactions to induced impulses. 
    The evaluation metrics are the same as in Tab.~\ref{tab:sota}. Best results are in boldface.
    }
    \label{tab:principled_action}
\end{table}

\begin{figure*}
    \centerline{    \includegraphics[width=\linewidth]{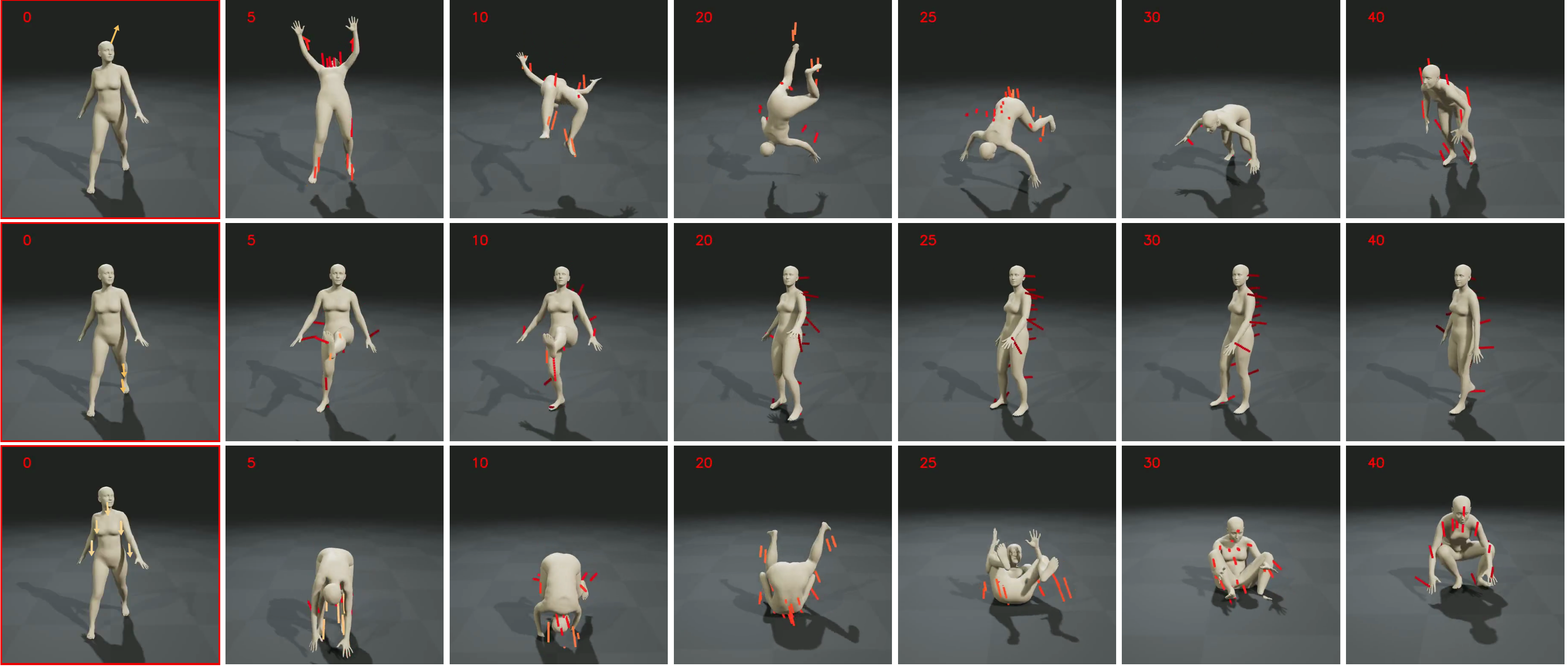}}
    \caption{Avatar reactions to induced impulses. Each row shows a motion. The \textcolor{orange}{pale orange arrows} in the initial frame denote the perturbed velocities, and \textcolor{red}{bars} in the future frames denote the generated joint locations and velocities. Darker color means smaller velocity norms.}
    \label{fig:principled_action_gen}
\end{figure*}

\paragraph{Reactions to induced impulses.}
Here, we perturb the velocities for subsets of the joints and quantify the resulting motions generated by the agent.
Depending on the initial velocities, we can make the agent generate kicks, forward runs, back flips, and forward rolls; see Sup.~Mat.~for the intuitive principles. 
With \model, we only perturb the first frame, but apply the velocities to all history frames of DiP and DART to improve their responsiveness.
To measure the motion realism, we generate a 3-second motion for each starting state, leading to 225 sequences per method, and compute ASR, ANCR, and ARD. 
In addition, we render 30 videos per action for a perceptual study that tests how well the models align with human intuitions; 
see the Sup.~Mat.~for details.
Table \ref{tab:principled_action} shows that \textit{InContext} outperforms the prior art.
Figure \ref{fig:principled_action_gen} visualizes some results; see \textbf{Supplementary Video}.
We see that the avatars have realistic reactions to the velocity interventions.

\paragraph{CBG-based control.} 
The CBG method proposed in Sec.~\ref{sec:method_cbg} provides control of the movement direction and speed of the agent. \textit{InContext} significantly outperforms the baselines on this task -- see Sup.~Mat.~for more details.

\begin{figure}
    \centering
    \includegraphics[width=\linewidth]{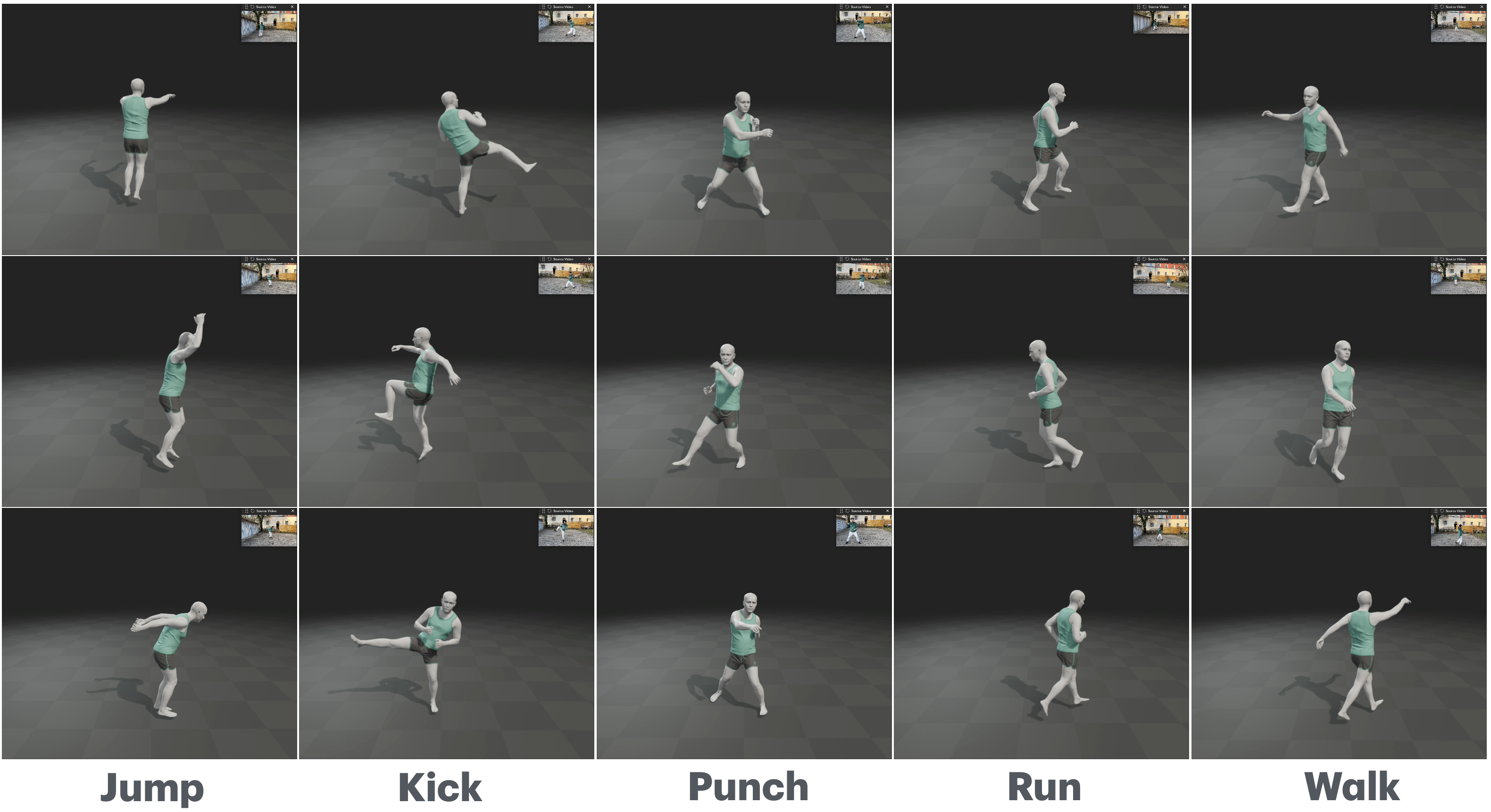}
    \caption{Snapshots of personalized mocap dataset, captured by MoCapade3.0 \cite{mocapade} from cellphone videos.}
    \label{fig:customized_action}
\end{figure}

\begin{table}
    \centering
    \footnotesize
    \begin{tabular}{lccc}
    \toprule
         \textit{Methods} & \textit{ASR} $\downarrow$  & \textit{ARD} $\times 10^{-3} \downarrow$ & R-precision $\uparrow$\\
         \midrule
         Finetuning & 0.253 & \textbf{0.060} & 0.87 \\
         OmniControlNet~\cite{xie2024omnicontrol} & 0.299  & {0.063} & 0.90 \\
         Ours & \textbf{0.215}  & {0.064} & \textbf{0.96} \\
         \bottomrule
    \end{tabular}
    \caption{Results of semantic action generation. R-precision indicates classification accuracy. See more results in Sup.~Mat. }
    \label{tab:adaptation_action}
\end{table}

\subsection{Evaluations on Model Adaptation}
We refer the reader to the Sup.~Mat.~for the model adaptation experiment details, which include \textbf{few-shot semantic action generation} and  \textbf{spatial target reaching}.
We evaluate two approaches:
1) We add the control embedding (i.e.~the embedding of $y$ in Fig.~\ref{fig:net_adaptation}) to the time embedding, and fine-tune the pretrained model directly. 
2) We implement the motion ControlNet proposed in~\cite{xie2024omnicontrol,zhong2024smoodi}, in which the control signal is fused to the input of the ControlNet only once.
We denote them as \textit{finetuning} and \textit{OmniControlNet}, respectively.
Overall, we find the ControlNet approach can adapt the \model base model to novel tasks efficiently and effectively.
Compared to \textit{OmniControlNet}, we find that our proposed adaptor enables the avatar to respond more quickly to the control signal.

For \textbf{few-shot semantic action generation}, we use Mocapade3.0 \cite{mocapade} to capture 5 classes of actions from video (Fig.~\ref{fig:customized_action}). Although the dataset is only about 3 minutes in total, it is sufficient to fine-tune the model.
Quantitative results are shown in Tab.~\ref{tab:adaptation_action}, while visualizations are in Fig.~\ref{fig:teaser}.
We observe that all these \model-based methods are effective.
The ControlNet approach allows adaptation of the base model to a small set of captured motions, maintaining the controllability and responsiveness of the original model, while being able to mimic the new data.

\section{Conclusion}
\label{sec:conclusion}
To create a motor model for interactive agents, we propose a new form of autoregressive diffusion model that generates perpetual, controllable, realistic, and responsive motions in real time.
Following recent advances in foundation models, our training paradigm has two phases, i.e. 1) unsupervised pretraining to learn body movements, and 2) adaptation to novel tasks.
Experimental results show that our model learns motions effectively, adapts to external velocity inputs, and can be efficiently adapted to few-shot personalized action generation and spatial target reaching.
We demonstrate the advantages with a real-time character animation system built in Unreal Engine.

\paragraph{Limitations and future work.}
First, while most motions exhibit perceptually physical realism, non-physical motions sometimes occur, e.g.~when getting up after falling, the avatar sometimes ``flies'' up.   
The method generalizes to falls that it has never seen, but more training data of this type would be helpful.
We clarify that we \textit{do not ensure physical realism}. 
Rather we generate 
motions that generally look physically realistic.
Second, \model has limited types of control inputs. Extending this to have more natural inputs through text or speech should be explored. For example, we can modify our ControlNet to fuse text and motion features, and fine-tune it on text-motion paired datasets, e.g. BABEL~\cite{BABEL:CVPR:2021} and HumanML3D~\cite{guo2022generating}.
Third, the agent has no awareness of its environment and does not know about obstacles. Learning to condition on the scene is future work.
Finally, the model could eventually replace traditional motion graphs and motion matching (cf.~\cite{Kovar2002,holden2015learning,holden2020learned}) in video games, however, despite running in real time (i.e. computation takes less than 0.5 second), the computational cost is still too much for this application.

\paragraph{Acknowledgments.} 
We sincerely thank Radu Alexandru Rosu, Yu Sun, and Aman Shenoy for the Rust-based SMPL-X rendering.
We are also heartily grateful to Guy Tevet for the help on CLoSD~\cite{tevet2024closd}. 
Moreover, we thank fruitful discussions with Muhammed Kocabas and Nikolas Hesse. 

\noindent\textbf{Disclosures.} 
While MJB is a co-founder and Chief Scientist at Meshcapade, his research in this project was performed solely at, and funded solely by, the Max Planck Society. This work was done while YF was an employee of Meshcapade.

{
    \small
    \bibliographystyle{ieeenat_fullname}
    \bibliography{main}
}

\clearpage

\setcounter{table}{0}
\setcounter{figure}{0}
\renewcommand{\thetable}{A\arabic{table}}
\renewcommand{\thefigure}{A\arabic{figure}}

\maketitlesupplementary

{
  \hypersetup{linkcolor=blue}
  \tableofcontents
}

\newpage

\section{More Discussions on \model }

\subsection{Motivation, Contribution, and Benefits Justification}

We aim to build a ``motor system'' for digital avatars enabling them to move perpetually and react promptly, like real humans.
This goes beyond existing game engines that animate characters with canned motions, increasing realism.
Inspired by recent advances in motion generation, \model generates perceptually realistic motions, generalizes across body shapes, supports efficient adaptation, and makes the avatar reactive to impulses. 
Our experiments and demos present its potential for game production, and show that physical simulation is not necessary to produce character animations that appear physically realistic.

\textit{The key novelty is the formulation that generates 0.5-second motion given a single initial state.}
This contrasts with prior work that generates a long future motion conditioned on a past motion.
Its benefits include reducing overfitting, making model training easier, and making the avatar reactive to impulses and classifier-based guidance.
Also, our ControlNet's advantages over existing ones are demonstrated in Sec.~\ref{sec:method:adaptation}, experiments, and the SupMat video.

\begin{figure*}
    \centering
    \includegraphics[width=\linewidth]{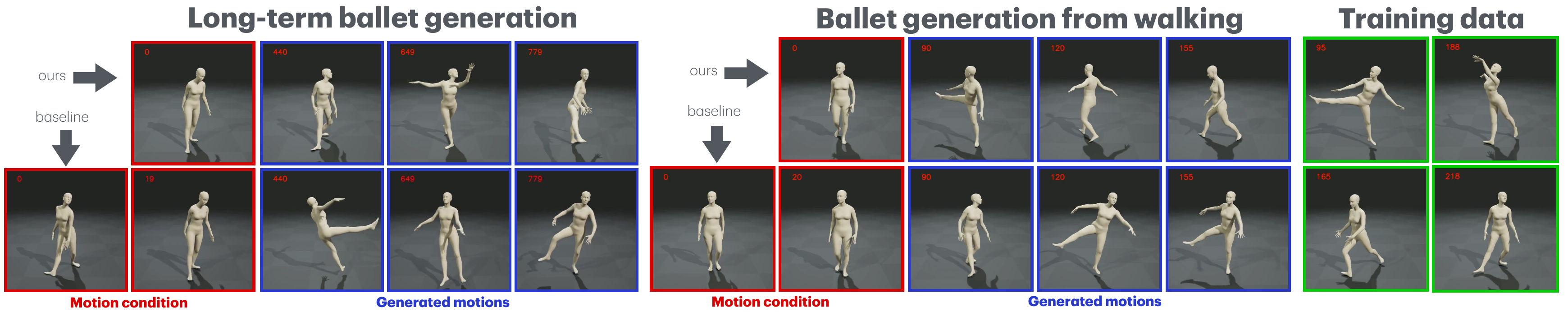}
    \caption{{Our half-second atomic action setting helps generalization and transition.} }
    \label{fig:reb_motion_length}
\end{figure*}

To better understand the benefits of our formulation,
we compare two identical settings except the motion length, where `ours' generates 15 frames given 1 frame, and `baseline' generates 40 frames given 20 frames.
We replace in-context with cross-attention to handle multi-frame conditioning.
Both models are successfully overfit to a ballet sequence with 229 frames, and they can reproduce the ballet conditioned on the start frames.
For testing, we first generate 780 future frames given the end frame(s) of that ballet sequence. 
We find `ours' produces ballet stably, whereas `baseline' gradually fails as time progresses. 
ASRs (Line475) are 0.08 (`ours') and 0.12 (`baseline').
Second, we generate 156 frames, with conditions from another walking sequence. 
We find `ours' produces natural transitions to ballet, whereas `baseline' produces severe artifacts. 
ASRs are 0.06 (`ours') vs 0.3 (`baseline').
These results indicate \textit{our setting makes the model more generalizable w.r.t.~motion length and semantics}.
Fig.~\ref{fig:reb_motion_length} shows some frames; videos are online on our website.

\subsection{Relations to Hierarchical Approaches}
Hierarchical approaches like CLoSD~\cite{tevet2024closd} and InsActor~\cite{ren2023insactor} first use a diffusion-based motion planner to generate a future motion, and then actuate an agent to track the generated motion in simulation.
\textit{We think this hierarchical setting is not suitable for interactive digital avatars.}
First, simulation tends to produce less natural looking motion than diffusion-based methods, which is shown by CLoSD's BPR in Tab.~1 and the \textbf{Supplementary Video} 6:33.
Second, simulation can bring extra computational costs and complicate the character animation workflow. 
PRIMAL only uses the diffusion model, but models much shorter motions. 
Unlike CLoSD (generates 40 given 20 frames) and InsActor (models more than 100 frames; see their code and SupMat), PRIMAL enables faster avatar responses.

\subsection{On Physical Realism Without Simulation}
We clarify that we do {\em not} ensure physical realism. 
Rather we generate \textit{perceptually realistic} motions that look physically realistic.
For example, the motion should not contain \textit{visible} artifacts such as foot skating or ground penetration.
Also, the generated motion should be human-like and aligned with user preferences. 
By learning purely from data, \model enables physics-like interactions with the agent (e.g.~pushing and pulling).
It generalizes to disruptions to the first frame that are out of distribution and responds naturally in many situations.
As mentioned in the conclusion (Sec.~\ref{sec:conclusion}), non-physical motions can occur, which could be overcome by using more training data, e.g.~of people getting up from the floor.

\section{More Method Demonstrations}

\subsection{Neural Architecture}
\label{sec:supp:architecture}

\begin{figure}
    \centering
    \includegraphics[width=0.85\linewidth]{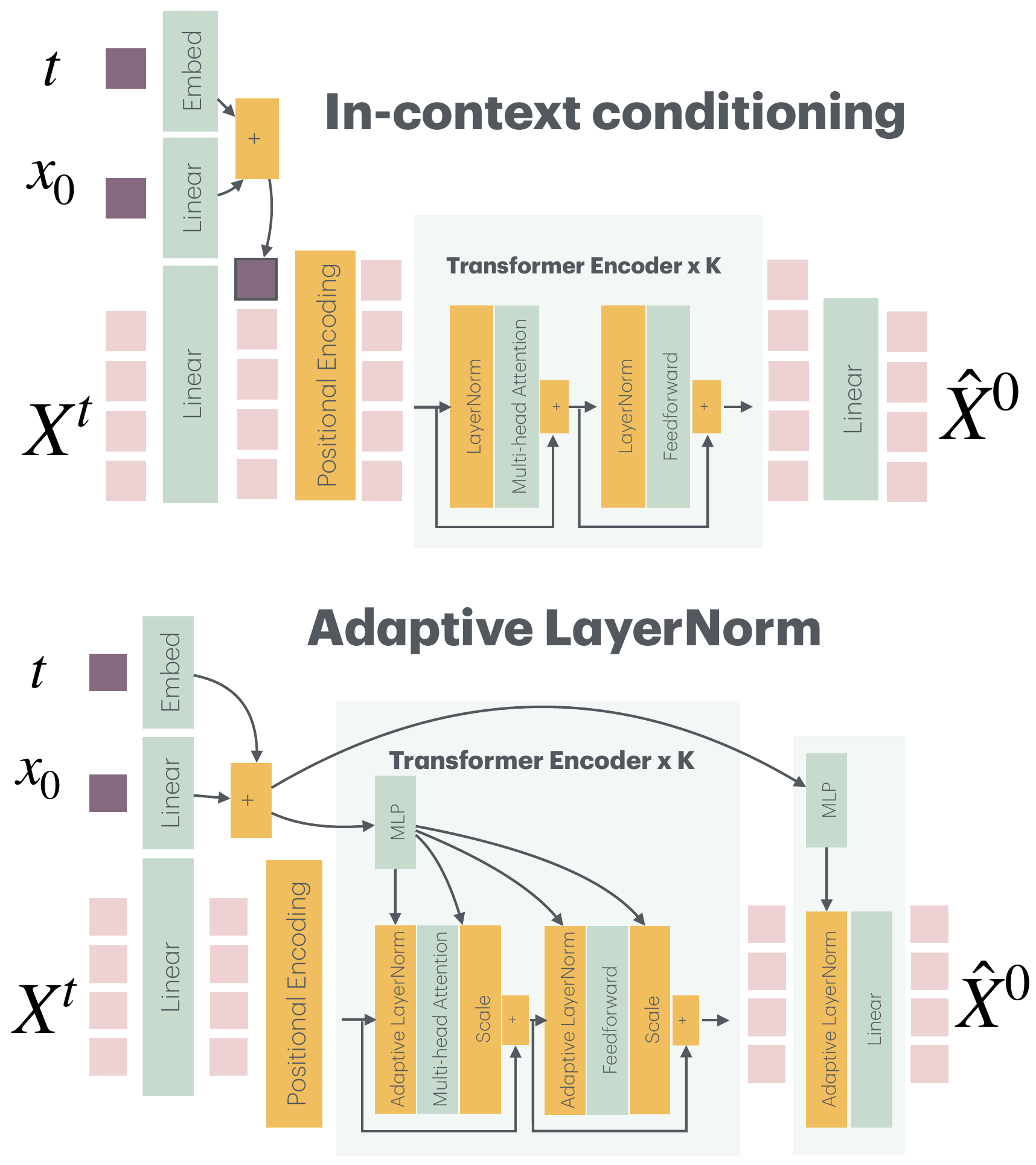}
    \caption{The diffusion network is formulated as $\hat{\bm{X}}^0 = G(t, {\bm{X}}^t, \bm{x}_0)$. 
    The tea-green layers contain trainable parameters and the orange blocks are non-learnable operations. 
    The pink squares denote the tokens at individual frames.}
    \label{fig:supp:base_models}
\end{figure}

We formulate our denoising network as $G(t, \mathbf{X}^t, \bm{x}_0)$ with transformers~\cite{vaswani2017attention,peebles2023scalable}, and investigate two architectures that differ in how the initial state is conditioned, as illustrated in Fig.~\ref{fig:supp:base_models}. 
The first architecture leverages in-context conditioning, in which the diffusion time embedding and the initial state embedding are first added and then concatenated with the embedding of the noisy motion segment.
We use SiLU~\cite{hendrycks2016gaussian,elfwing2018sigmoid,swish} in the feed-forward layers.
The second architecture follows the DiT network~\cite{peebles2023scalable} with adaptive layer normalization~\cite{perez2018film}.

\subsection{Inference}
\label{sec:supp:inference}

\subsubsection{The Inference Algorithm}

Our motion model learns the distribution of $p({\bm X} | {\bm x}_0)$, following the notations in Sec.~\ref{sec:method}.
Given an initial state, the model generates a motion segment starting with this initial state, and  works recursively to produce an arbitrarily long motion sequence.
This generation process can be refined and controlled on the fly.
The overall inference approach is summarized in Alg.~\ref{alg:inference}.

\begin{algorithm}[t]
\footnotesize
\SetAlgoLined
\KwResult{A new motion segment}
 \textbf{Init}: An initial state $\bm{x}_0$, guidance losses (optional)\;
 \textbf{Step 1}: canonicalize the initial state\;
 \Indp
     1.1 transform the initial state to body-centric coordinates\;
     1.2 joint re-projection (optional)\;
 \Indm
 \textbf{Step 2}: condition embedding\;
 \textbf{Step 3}: generation process\;
    \Indp
    3.1 DDPM reverse diffusion process\;
    \Indm
     \Indp
     \For{each denoising step}{
     
      predict the clean motion segment\;
      derive a Gaussian distribution to sample\;
      compute the gradients of the guidance losses (optional)\;
      update the previous Gaussian distribution (optional)\;
      sample and get a less noisy motion segment\;
      
     }
     \Indm
     \Indp
    3.2 snapping to ground (optional)\;
    3.3 inertialization blending (optional)\;
    \Indm
 \textbf{Step 4}: transform the motion segment back to world coordinates\;
     
\caption{The algorithm for generating a motion segment. Long-term motions can be generated by applying this approach repeatedly.  }
\label{alg:inference}
\end{algorithm}

\subsubsection{Test-time Processing}
\label{sec:supp:ttprocessing}

To further improve the motion quality, we propose the following three processing steps before or after generating each motion segment. 
Although they can be used at each individual DDPM denoising step, this introduces extra computational costs, and we did not observe advantages. 

\paragraph{Joint re-projection.}
As time progresses, the predicted joints can drift from the SMPL-X body. 
Inspired by~\cite{zhang2022wanderings}, in each initial state we re-compute the joint locations based on the predicted SMPL-X parameters, and replace the predicted joint locations with these re-projected values.  
We then use this modified initial state to generate the current motion segment.

\paragraph{Snapping to the ground.}
A key aspect of motion realism is good foot-ground contact.
When the model jumps, it may not fully return to the ground, resulting in foot sliding in future frames.
To address this, we determine the lowest body joint over a 0.5 second motion segment. 
Given our hang-time assumption, in this time frame, some joint must be in contact with the ground.  
We then translate the entire segment along the vertical direction to ``snap'' the body on the ground.
Since the motion segment spans roughly the hanging time of jump, this does not degrade the avatar's ability to jump.

\paragraph{Inertialization blending.}
In some cases, especially after snapping, discontinuities between the initial state and the generated motion segment can appear, leading to jittering artifacts.
Therefore, we perform inertialization~\cite{inertialization,inertialization_d} to smooth the transitions. 

Perceptually, these three processing steps reliably improve the motion quality. 
Their quantitative evaluations are referred to~\cref{tab:ablation_base}.

\subsubsection{Avatar Reactions to Induced Impulses}

A sign of the emergence of physical effects is that the avatar can react to external impulses promptly and naturally.
In this case, we can control the avatar with impulses to generate certain actions in a principled manner. 
For example, we can generate a kick by giving an upward impulse to the joints on the leg, or generate a run by pushing the avatars from the back.

In this work, we induce impulses and forces to interact with the avatar, by intervening the initial velocities before generating a new motion segment.
Based on our intuitions, we find some actions can be reliably generated, shown in Tab.~\ref{tab:five_actions}. Their experimental results are presented in Sec.~\ref{sec:exp:principled}.  Automatic solutions to discovering action principles are interesting to explore in the future.

\begin{table}
    \centering
    \footnotesize
    \begin{tabular}{lcc}
    \toprule
         \textit{action} & \textit{perturbed joints} & \textit{perturbing velocities} \\
         \midrule
         kick, left leg & left knee, left ankle & $(0,0,0.5)$ \\
         kick, right leg & left knee, left ankle & $(0,0,0.5)$ \\
         forward run & pelvis, neck, shoulders & $(0,0,1)$ \\
         back flip & head & $(0,0,-1)$ \\
         forward roll & head, shoulders, elbows & $(0,-0.5,0.5)$\\
         \bottomrule
    \end{tabular}
    \caption{Action generation based on intuitions and straightforward principles. These perturbing velocities are in the body-centric coordinate, and are added to the mentioned joints in the initial states.
    }
    \label{tab:five_actions}
\end{table}

\section{More Experimental Details} 
\label{sec:supp:exp}

\subsection{Baselines for Motion Realism Evaluation}

We compare our methods with two groups of baselines.
The \textit{off-the-shelf} group includes the pretrained DiP model~\cite{tevet2024closd} for online text-to-motion generation. 
We run their released code to produce 240 sequences of SMPL joint locations at 20fps, based on the test set of HumanML3D~\cite{guo2022generating}. 
For fair comparison, we upsample the generated joint locations to 30fps via cubic interpolation and trim them to 240 frames to compute the quantitative metrics. 
For the perceptual study, we fit gender-neutral SMPL-X bodies to the joints via optimization, and render the videos with the same pipeline as ours.
We also run the full version of CLoSD~\cite{tevet2024closd} based on their given test set, and obtain the SMPL joint locations. Then we process and evaluate these results as for the pretrained DiP.
In addition, we run the off-the-shelf MotionLCM~\cite{dai2024motionlcm}, which is a SOTA text-to-motion approach that runs in realtime. 
Since its base model focuses on text-to-motion and is not autoregressive, we use its default motion lengths for testing to ensure the motion quality. Likewise, we upsample the generated joint trajectories to 30fps before computing the metrics.

The \textit{``our implemented''} group in Tab.~\ref{tab:sota} contains the DiP and the DART~\cite{Zhao_DART_2024} diffusion model.
Their motion representations and pretraining strategies are identical to ours.
To focus on learning the body motion, we remove their text encoders but keep their key designs.
Specifically, our DiP version conditions the time embedding via cross attention. 
Based on 20 frames in the past, it generates 40 future frames.
Our DART diffusion model generates 8 future frames based on 2 past frames, suggested in~\cite{Zhao_DART_2024}. Neither VAE tokenizers nor scheduled sampling are used. 
Their training losses are also based on Eq.~\eqref{eq:our_loss}, except that the reconstructed past motions are ignored, following their original settings. 
During inference, the three test-time processing steps in Sec.~\ref{sec:method_inference} (or Sec.~\ref{sec:supp:ttprocessing}) are also applied, in order to reduce artifacts such as ground interpenetration and jittering.

\begin{table}
    \centering
    \footnotesize
    \begin{tabular}{lccccc}
    \toprule
         \textit{Methods} & \textit{ASR} $\downarrow$  & \textit{ARD} $\times 10^{-3} \downarrow$ & err. vel. $\downarrow$ & err. dir. $\downarrow$\\
         \midrule
         DART diffusion & 0.118 & 3.622 & 2.532 & 1.022 \\
         DiP & 0.223 & 4.235 & 1.341 & 0.596 \\
         \midrule
         InContext-8f &  0.139 & 0.773 & 2.519 & 0.262 \\
         InContext & \textbf{0.109} & \textbf{0.252} & \textbf{1.040} & \textbf{0.140} \\
         \bottomrule
    \end{tabular}
    \caption{The results of the CBG-based control. `err. vel.' and `err. dir.' denote the averaged L2 distances between the generated velocity/facing direction to their targets. }
    \label{tab:cbg_control}
\end{table}

\begin{figure*}
    \centering
    \includegraphics[width=0.9\linewidth]{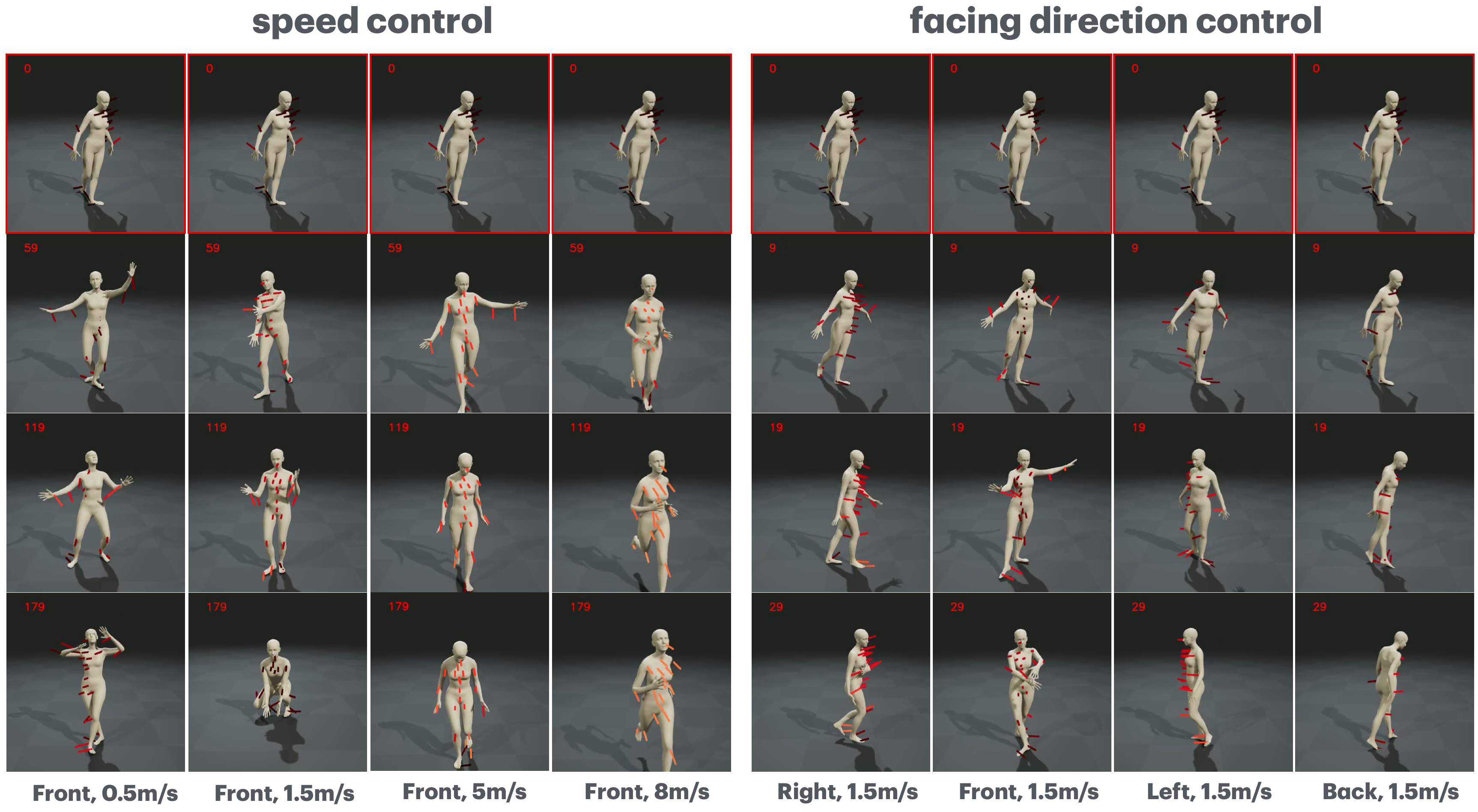}
    \caption{Illustrations of CBG-based control. Each column visualizes a motion sequence. The texts at the bottom denote the targets to achieve. The \textcolor{red}{bars} show the generated joint velocities, with darker/brighter color denoting smaller/larger velocity norms. }
    \label{fig:cbg_control}
\end{figure*}

\subsection{{CBG}-based Control of Speed and Direction} 
We evaluate the CBG method proposed in Sec.~\ref{sec:method_cbg}, which is based on intuition of the movement speed and the direction.
The guidance weights of $\mathcal{L}_{\mathit{move}}$ and $\mathcal{L}_{\mathit{facing}}$ (see \cref{sec:method_cbg}) are set to 50 and 25, respectively.
We specify 4 target directions to move, i.e. (forward, left, right, back), and each direction has 2 target speeds of (1m/s, 4m/s).
The target directions are applied to both moving and facing.
Based on each individual initial state and target, we generate a 3-second motion, leading to 360 generated motions for each method.
To evaluate the performance, we compute the L2 distance between the generated values and the target values, as well as the motion realism metrics.
Considering the transition phase, we compute these L2 distances after 1 second.
The results are shown in Tab.~\ref{tab:cbg_control}. 
We find that DART and \textit{InContext-8f} are not responsive to the guidance on the mean velocity, probably because the average velocity in 0.25 second is highly ambiguous.
The DiP model has a similar problem. Also, it tends to align the semantics between the past and the future motion, and hence increases the risk of mismatch between the motion history and the goal, leading to unrealistic movements.
Fig.~\ref{fig:cbg_control} illustrates our CBG control process. 
We can see that the avatar can react to the facing direction guidance within one second.
See Supplemental Video.

\subsection{Model Adaptation}

We adapt \model-\textit{InContext} to the tasks of semantic action generation and spatial target reaching. 
We evaluate two approaches:
1) We add the control embedding (i.e.~the embedding of $y$ in Fig.~\ref{fig:net_adaptation}) to the time embedding, and fine-tune the pretrained model directly. This is also called as \textit{injection} in some literature.
2) We implement the motion ControlNet proposed in~\cite{xie2024omnicontrol,zhong2024smoodi}, in which the control signal is fused to the input of the ControlNet only once.
An exponential moving average is also applied in the adaptation phase. 
We denote them as \textit{finetuning} and \textit{OmniControlNet}, respectively.

\subsubsection{Personalized Action Generation}

We use Mocapade3.0 \cite{mocapade} to capture 5 stylized action classes from video at 30fps (Fig.~\ref{fig:customized_action}).
Each class contains about 5-7 sequences of 5 seconds. 
We adapt the base model to this dataset using AdamW~\cite{kingma2014adam,loshchilov2017fixing} with a learning rate of 0.0001, and set the batch size to 16. The training terminates after 1000 epochs, leading to 19K iterations.
In addition to test our mentioned baselines, we also train our model, i.e. the one used for \textit{finetuning}, on this dataset from scratch.

To evaluate the action conditioning, we train an action label and motion feature extractor via contrastive learning~\cite{chopra2005learning}, following the evaluation metrics of~\cite{chopra2005learning}.
The motion encoder is a transformer with 4 blocks, 256 latent dimensions, and 8 heads. 
The action encoder is a single embedding layer.
These encoders are trained based on our customized action data.
Since our task is few-shot adaptation instead of large-scale generalization, we compute the top-1 R-precision (equivalent to classification accuracy here), and ignore the FID metric that is sensitive to the data size.
We also measure the motion realism as before.

We draw 45 initial states from SFU~\cite{mocap_sfu}, and generate a 3-second motion for each individual action. 
Results are shown in Tab.~\ref{tab:supp:adaptation_action}. 
We can see that all methods are effective. Training from scratch and our adaptation method perform best to reproduce the personalized actions, whereas our method is more realistic. 
Another advantage of the ControlNet-based adaptation is that it supports CFG to manipulate the intensity of the motion style.
See the \textbf{Supplementary Video} for more details.

\begin{table}
    \centering
    \footnotesize
    \begin{tabular}{lccc}
    \toprule
         \textit{Methods} & \textit{ASR} $\downarrow$  & \textit{ARD} $\times 10^{-3} \downarrow$ & R-prec. $\uparrow$\\
         \midrule
         scratch training & 0.314  & 0.059 & \textbf{0.96} \\
         finetuning & 0.253 & \textbf{0.060} & 0.87 \\
         OmniControlNet~\cite{xie2024omnicontrol} & 0.299  & {0.063} & 0.90 \\
         ours & \textbf{0.215}  & {0.064} & \textbf{0.96} \\
         \bottomrule
    \end{tabular}
    \caption{Results of few-shot adaptation to action generation. All ANCR results are very close to 1.}
    \label{tab:supp:adaptation_action}
\end{table}

\subsubsection{Spatial Target Reaching}

We leverage the same training datasets with \textit{InContext} of \model. 
The learning rate is fixed to 0.0001, and the batch size is set to 128. The adaptation phase terminates after 100 epochs or 6.3K iterations.

We draw 45 initial states from SFU as in Sec.~\ref{sec:exp:principled}.
For each initial state, we specify 8 spatial targets w.r.t. the initial pelvis location, i.e. $(\pm{1.5},0), (0,\pm{1.5}),(\pm{1},0), (0,\pm{1})$ in meters.
Besides the above two baselines, we run the off-the-shelf DiP model~\cite{tevet2024closd} conditioned on the above pelvis goal locations. Each pelvis target also has 45 different motions. 
We zero mask the text embedding to let the model focus on the spatial target conditioning.
We use this DiP model to produce sequences of joint locations, and upsample them to 30 fps via cubic interpolation.

For each target and each initial state, we generate a 2-second motion, leading to 360 sequences per method.
We compute the minimal distance between the generated pelvis 2D trajectory and the target location for evaluation, as well as the motion realism metrics.

The results are shown in Tab.~\ref{tab:adaptation_goal}.
We can find that directly finetuning is an effective approach, and has comparable performance with the two ControlNet-based methods. 
Compared to OmniControlNet, our adaptation method leads to a smaller distance error but slightly worse motion realism, indicating that the control signal has more impact.

\begin{table}
    \centering
    \footnotesize
    \begin{tabular}{lccc}
    \toprule
         \textit{Methods} & \textit{ASR} $\downarrow$  & \textit{ARD} $\times 10^{-3}$ $\downarrow$ & err. dist. $\downarrow$ \\
         \midrule
         DiP-goal~\cite{tevet2024closd} & {0.3}  & - & 0.177 \\
         \midrule
         finetuning & {0.259} & {0.399} & {0.071} \\
         OmniControlNet~\cite{xie2024omnicontrol} & \textbf{0.249} & \textbf{0.366} & 0.106 \\
         ours & {0.338} & 0.393 & \textbf{0.031}\\
         \bottomrule
    \end{tabular}
    \caption{Results of adaptation to spatial target reaching. The ANCR values of all methods are close to 1. Since SMPL-X fitting is not applied to DiP results, its ARD value is not available. }
    \label{tab:adaptation_goal}
\end{table}

\begin{table*}
    \centering
    \begin{tabular}{lcccccc}
    \toprule
         & \multicolumn{3}{c}{HumanEva} & \multicolumn{3}{c}{SFU} \\
           \cmidrule(lr){2-4}  \cmidrule(lr){5-7}
         \textit{Methods} & \textit{ASR} $\downarrow$ & \textit{ANCR}  $\uparrow$ & \textit{ARD} $\times 10^{-3} \downarrow$  & \textit{ASR} $\downarrow$ & \textit{ANCR}  $\uparrow$  & \textit{ARD} $\times 10^{-3} \downarrow$ \\
         \midrule
         InContext & \textbf{0.017} & 1.0 & 0.059 & \textbf{0.027} & 1.0 & \textbf{0.081} \\
         AdaLN & 0.031 & 1.0 & 0.109 & 0.039 & 1.0 & 0.167 \\
         InContext w/ noise & 0.020 & 1.0 & 0.089 & \textbf{0.022} & 1.0 & 0.098 \\
         InContext, ReLU & 0.028 & 1.0 & {0.058} & 0.042 & 1.0 & 0.101 \\
         InContext, ReLU, separate tokens~\cite{camdm} & 0.037 & 1.0 & \textbf{0.053} & 0.040 & 1.0 & 0.082\\
         InContext, 512d & 0.023 & 1.0 & 0.072 & 0.027 & 1.0 & 0.105 \\
         InContext, 8 frames & 0.036 & 1.0 & 0.122 & 0.073 & 1.0 & 0.096 \\
         \midrule
         InContext &  &  &  & & &   \\
         no proc. & 0.022 & 1.0 & \textbf{0.055}  & 0.037 & 1.0 & 0.082 \\
         + reproject & 0.034 & 1.0 & \textbf{0.055} & 0.042 & 1.0 & \textbf{0.078} \\
         + inertialize & \textbf{0.015} & 1.0 & 0.069 & 0.028 & 1.0 & 0.089 \\
         + reproject + inertialize & 0.017 & 1.0 & 0.059 & \textbf{0.027} & 1.0 & 0.081 \\
         \bottomrule
    \end{tabular}
    \caption{Ablation studies on the base models. The initial state and every generated motion segments are snapped to the ground. The first part compares different model architectures. The second part shows the influences of the inertialization and joint re-projection. 
    }
    \label{tab:ablation_base}
\end{table*}

\subsection{Ablation Studies of \model}
Tab.~\ref{tab:ablation_base} shows the ablation study results of our \model methods, in which different model instances are separately trained from scratch.
In the version of \textit{separate tokens}, we follow CAMDM~\cite{camdm} to concatenate the time embedding and the initial state embedding in the context dimension instead of adding them, in which the activation function is ReLU.
Since \textit{InContext} consistently outperforms \textit{AdaLN}, our ablation studies are mainly based on \textit{InContext}.

Inspired by~\cite{chen2025diffusion,valevski2024diffusionmodelsrealtimegame}, we add Gaussian noise with 0.01 standard error to the joint locations and velocities, but no advantages are observed in our trials. 
Additionally, the following approaches could not bring consistent better performances: 1) replacing SiLU with ReLU; 2) lifting the attention latent dimension from 256 to 512; 3) predicting 8 frames rather than 15 frames; 4) the \textit{separate tokens} scheme proposed in~\cite{camdm}.

We further investigate the influences of our test-time processing approaches.
Joint re-projection can slightly increase the skating ratio and reduce the distances to the AMASS pose manifold.
Inertialization can reduce skating and eliminate jittering artifacts effectively.
Leveraging both of them is a good practical balance.

\begin{table*}
    \centering
    \begin{tabular}{lccccc}
    \toprule
         \textit{Setting} & Step 1  & Step 2 & Step 3 & Step 4 & Total\\
         \midrule
         50 denoise steps & 8.195  & 0.138 & 152.745 & 3.695 & 164.773 \\
         {50 denoise steps +  test-time proc.} & 8.909  & 0.129 & 152.635 & 3.425 & 165.099 \\
         {50 denoise steps +  test-time proc. + CBG} & 9.348  & 0.138 & 192.789 & 3.607 & 205.881 \\
         {50 denoise steps +  test-time proc. + CBG + ControlNet} & 10.687  & 0.723 & 349.729 & 3.540 & 364.679 \\
         {10 denoise steps +  test-time proc. + CBG + ControlNet} & 10.479  & 0.725 & 71.660 & 3.465 & 86.329 \\
         \bottomrule
    \end{tabular}
    \caption{Results of runtime analysis to generate a 0.5-second motion segment. The numbers are in millisecond. Step 1-4 are according to our inference algorithm~\ref{alg:inference}.}
    \label{tab:runtime}
\end{table*}

\subsection{Default Implementations}
\label{sec:exp:implementation}
By default, in \model the transformer has 10 blocks and the dropout ratio is 0.1. 
In each transformer block, the self-attention layer has the latent dimension 256 and 8 heads. 
The feed-forward intermediate layer has 2048 dimensions, and its activation function is SiLU~\cite{hendrycks2016gaussian,elfwing2018sigmoid,swish}.
The DDPM process has 50 steps during training, following~\cite{camdm,tevet2024closd,Zhao_DART_2024,shi2024amdm}. 
We use AMASS~\cite{AMASS:ICCV:2019} to pretrain our diffusion models~\footnote{Our training sets are ACCAD~\cite{AMASS_ACCAD}, BMLmovi~\cite{AMASS_BMLmovi}, BMLrub~\cite{AMASS_BMLrub}, CMU~\cite{AMASS_CMU}, DFaust~\cite{AMASS_DFaust}, Eyes Japan Dataset~\cite{AMASS_EyesJapanDataset}, HDM05~\cite{AMASS_HDM05}, PosePrior~\cite{AMASS_PosePrior}, SOMA~\cite{AMASS_SOMA}, MoSh~\cite{AMASS_MoSh}, SSM.}. 
All sequences are downsampled to 30fps first. 
During training, 15-frame motion segments are randomly extracted from a batch of sequences at each iteration.
We use AdamW~\cite{kingma2014adam,loshchilov2017fixing} optimizer and keep the learning rate at 0.0001. 
The batch size is fixed to $512$, and the training terminates after 30K epochs, leading to 480K iterations in total. Exponential moving average (EMA) with decay $0.999$ starts to apply after the 1500-th epoch.
We use a single NVIDIA H100 GPU for training, taking 3-4 days to pretrain a specific version of \model. 
During inference, we use the checkpoints of EMA. 
By default, the reverse diffusion process takes 50 steps, and the snapping-to-ground operation (see Sec.~\ref{sec:supp:ttprocessing}) is used.

\subsection{Runtime Analysis}

We implement our methods with a single NVIDIA RTX 6000 GPU to measure the runtime of inference (see Algo.~\ref{alg:inference}).
We generate a 1200-frame motion sequence by generating motion segments recursively, then we compute the average runtime of each individual step. 
The results are shown in Tab.~\ref{tab:runtime}.
We can see that all settings can produce a motion segment within 0.5 second, allowing us to stream the generated frames to the frontend (e.g. Unreal Engine) in real time.  
Additionally, our test-time processing is efficient. Enabling the CBG-based control brings trivial computation cost, due to their analytical gradients.
Our ControlNet-based adaptor can largely slow down the diffusion, which can be overcome by reducing the denoising steps. 
In our trials, we find using 10 denoising steps does not bring visibly inferior results.

\subsection{Perceptual User Studies}

\begin{figure}
    \centering
    \includegraphics[width=\linewidth]{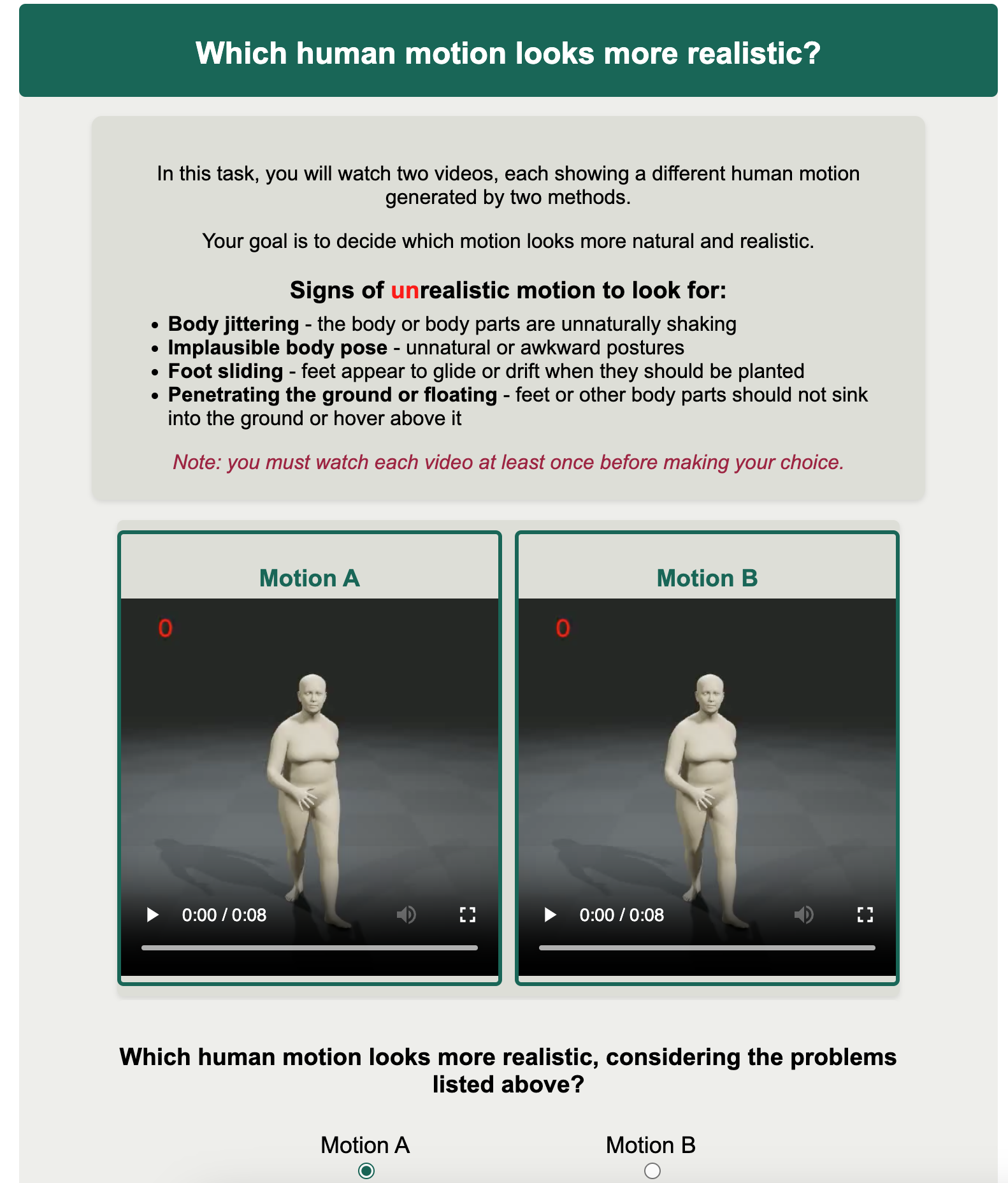}
    \caption{
        Layout of the motion realism perceptual study. Participants are presented with two videos of rendered motions and need to pick the more realistic one, according to the instructions.
    }
    \label{fig:mturk_layout_realism}
\end{figure}

\begin{figure}
    \centering
    \includegraphics[width=\linewidth]{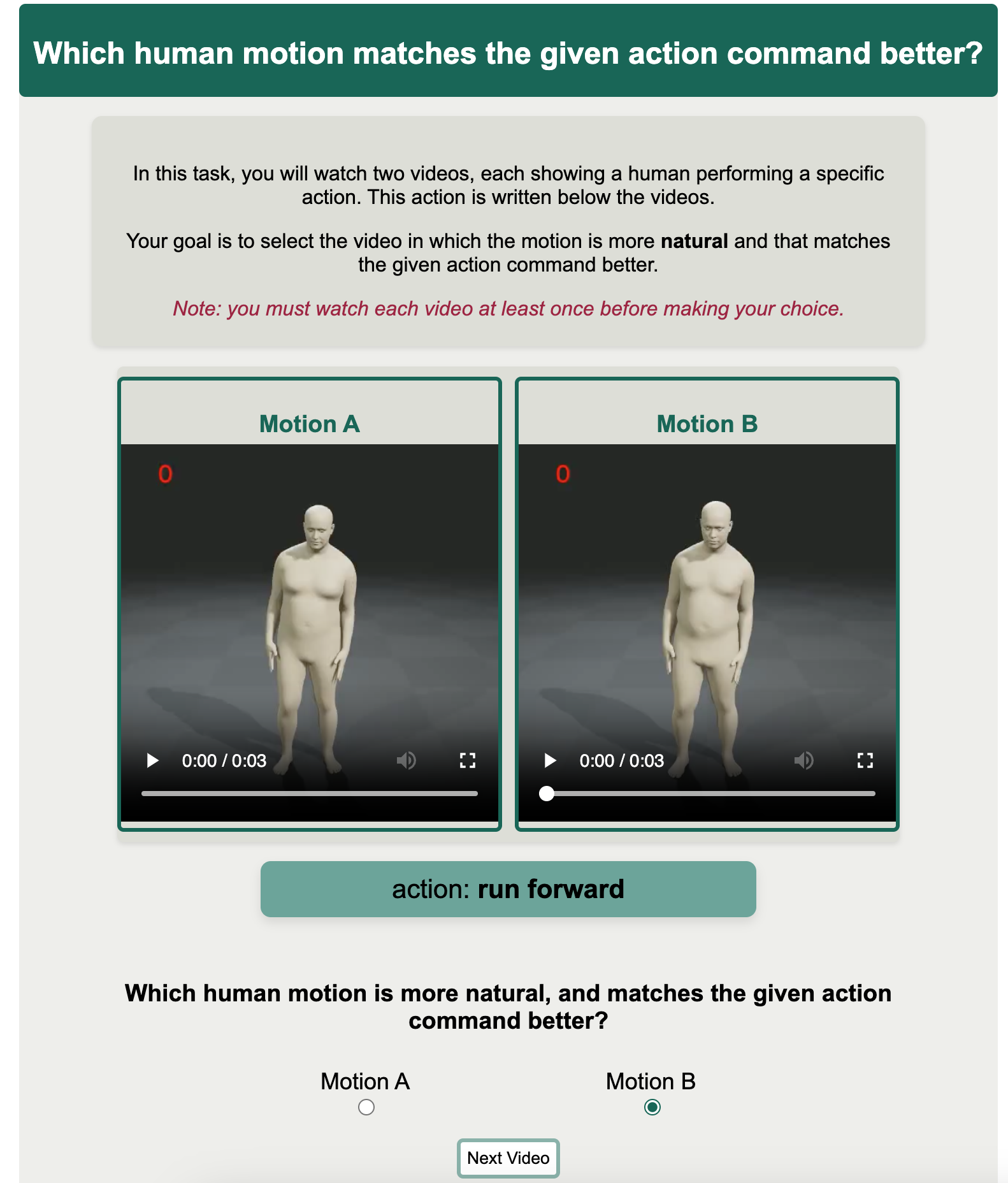}
    \caption{
        Layout of the action generation perceptual study. Below the instructions, participants are presented with two videos of rendered motions and a common action label. 
    }
    \label{fig:mturk_layout_actiongen}
\end{figure}

We conduct perceptual studies on the Amazon Mechanical Turk platform for both of our experiments mentioned in Sec.~\ref{sec:exp:motion_realism} and Sec.~\ref{sec:exp:principled}.
In summary, we present users with paired results—one from our method and one from a baseline method. Users choose the result they prefer, and we report the percentage of cases in which the baseline is preferred.
The layouts of the two perceptual studies are shown in \cref{fig:mturk_layout_realism,fig:mturk_layout_actiongen}.

Besides only allowing experienced and highly rated participants, we take several precautions in our study protocol to ensure reliable results.
Each assignment contains 57 comparisons, i.e. pairs of videos. The first 3 are intended as warm-up tasks, and the answers to these are not considered during evaluation. There are 4 catch trials too. These are intentionally obvious comparisons that help us identify participants who are providing random inputs. We discard all submissions where even a single one of the four catch trials is failed: 6 out of a total of 55 assignments.

To eliminate bias, the order of the other 50 actual comparisons is shuffled within an assignment, and the two sides of each comparison are randomly swapped too. All methods use the same rendering pipeline.

\end{document}